\documentclass[authoryear,preprint,review,12pt]{elsarticle}

\usepackage{amssymb}
\usepackage[dvipsnames,svgnames,x11names]{xcolor}
\usepackage[markup=underlined]{changes}
\usepackage{amsmath}
\usepackage{makecell}
\usepackage{tabularx}
\usepackage{multirow}

\usepackage{amsthm}

\journal{Environmental Modelling \& Software}

\begin{document}

\begin{frontmatter}



\title{Prediction of Sea Ice Velocity and Concentration in the Arctic Ocean using Physics-informed Neural Network}


\author[a,b,c]{Younghyun Koo}
\author[a,b]{Maryam Rahnemoonfar}

\affiliation[a]{organization={Department of Computer Science and Engineering, Lehigh University},
            addressline={27 Memorial Dr W}, 
            city={Bethlehem},
            postcode={18015}, 
            state={PA},
            country={USA}}

\affiliation[b]{organization={Department of Civil and Environmental Engineering, Lehigh University},
    addressline={27 Memorial Dr W}, 
    city={Bethlehem},
    postcode={18015}, 
    state={PA},
    country={USA}}

\affiliation[c]{organization={National Snow and Ice Data Center, Cooperative Institute for Research in Environmental Sciences, University of Colorado Boulder},
            addressline={1540 30th St},
            city={Boulder},
            postcode={80303}, 
            state={CO},
            country={USA}}

\begin{abstract}
As an increasing amount of remote sensing data becomes available in the Arctic Ocean, data-driven machine learning (ML) techniques are becoming widely used to predict sea ice velocity (SIV) and sea ice concentration (SIC). However, fully data-driven ML models have limitations in generalizability and physical consistency due to their excessive reliance on the quantity and quality of training data. In particular, as Arctic sea ice entered a new phase with thinner ice and accelerated melting, there is a possibility that an ML model trained with historical sea ice data cannot fully represent the dynamically changing sea ice conditions in the future. In this study, we develop physics-informed neural network (PINN) strategies to integrate physical knowledge of sea ice into the ML model. Based on the Hierarchical Information-sharing U-net (HIS-Unet) architecture, we incorporate the physics loss function and the activation function to produce physically plausible SIV and SIC outputs. Our PINN model outperforms the fully data-driven model in the daily predictions of SIV and SIC, even when trained with a small number of samples. The PINN approach particularly improves SIC predictions in melting and early freezing seasons and near fast-moving ice regions.


\end{abstract}




\end{frontmatter}


\section{Introduction}
\label{introduction}

Dramatic declines in the Arctic sea ice extent (SIE) and sea ice thickness (SIT) have been observed for the last few decades along with a warming climate resulting from the anthropogenic CO2 emissions \citep{Notz2016}. Arctic SIE has been reduced by more than 50,000 $\text{km}^\text{2}$/year, and at the same time, Arctic SIT has decreased by more than 2 m in average, with more than a 50 \% SIT reduction in thick multi-year ice \citep{Kwok2018}. With such dramatic changes in SIE and SIT, the thermodynamic and dynamic conditions of Arctic sea ice have entered a new phase \citep{Wang2022_new, Stroeve2018}. Given that sea ice plays a critical role in the physical conditions of the atmosphere and ocean (e.g., heat exchange, salinity, etc.) \citep{haumann2016sea, Hirst1999, mcphee1992turbulent, BUDIKOVA2009}, understanding sea ice dynamics is essential for understanding the future of the Arctic Ocean and global climate.


A traditional way to predict the sea ice dynamics in the Arctic Ocean is using numerical models based on physics laws and equations that describe sea ice in terms of mass and momentum balances, as well as its interaction with the atmosphere and ocean \citep{Stark2008, Fritzner2019, Hibler1979, Parkinson1979}. Although physics models have successfully explained general sea ice behaviors by assuming sea ice as viscous-plastic or elastic-viscous-plastic material \citep{Hunke1997, Hibler1979}, several challenges make it difficult to improve their fidelity and efficiency. First, since countless atmospheric, oceanic, and even biological variables are involved in sea ice dynamics, established physics equations have limitations in explaining every detailed mechanism of sea ice dynamics. If the sea ice dynamics are partially governed by any unknown mechanisms, physics models may not agree perfectly with the real observations. Second, uncertainties in the complicated parameterization of sea ice physical conditions, including initial and boundary conditions, can propagate through the model predictions \citep{Blockley2020, Hunke2010}. Since real observations are not always available, physics models are often based on physical assumptions. Considering that numerical models are sensitive to initial conditions and physical assumptions, they can result in significant uncertainties and discrepancies with real observations \citep{Wrigglesworth2015}. Third, solving coupled and non-linear partial differential equations (PDEs) is computationally intensive and expensive. Although adding more complexity to a model or increasing model resolution can enhance the model's accuracy, it exponentially increases computational costs \citep{Hunke2010, Li2023}. Hence, this computational demand inhibits improving the model performance for higher spatial and temporal resolutions.

Recently, machine learning (ML) techniques have emerged to predict sea ice behaviors with fewer complexities and a lower computational cost than traditional numerical models. ML approaches have several benefits compared to numerical models. First, complex PDEs or parameterizations are not required for ML models because they need only observational data to be trained. ML models can learn complex non-linear patterns or relationships between multiple atmospheric and oceanic variables directly from observational data without any explicit knowledge of physical processes. Second, ML models can be easily updated and trained continuously with new observational data, and this data integration can make the model prediction more accurate and reliable. Finally, ML models require less computational cost than numerical models. While numerical models often require intensive computing resources (e.g., high-performance computing) to solve complex PDEs, ML models can be implemented relatively fast by leveraging the parallel processing ability of graph processing units (GPUs).

Based on such advantages, various machine learning models have been developed to predict sea ice concentration (SIC) and sea ice velocity (SIV), most of which rely on convolutional neural networks (CNN) \citep{Ren2021, Liu2021_daily, Yan2018, Hoffman2023, Petrou2019}. As a deep learning algorithm specialized for image data, CNN propagates the input data through convolutional layers with multiple filters and extracts spatial patterns and features. Based on the ability of CNN to learn complex features in image datasets, CNNs have succeeded in various applications in environmental modeling \citep{WANG2024_EMS_CNN, JIANG2023_EMS_CNN, SADEGHI2020_EMS_CNN}. Since the SIC and SIV data for the entire Arctic Ocean are often provided from remote sensing data sources as grid formats, CNN has advantages in exploiting the spatial variability in those sea ice data.

However, the fidelity of fully data-driven ML models is highly dependent on the quantity and quality of training datasets, which leads to limitations in further improving the model performance. First and foremost, data-driven ML models require a large amount of data to ensure generalizability. If training samples are insufficient or their distribution is biased, the model can be overfitted and lack generalizability to out-of-training cases. Second, ML models are considered \emph{black-box} models and lack inherent interpretability in how the final predictions are derived. This lack of inherent interpretability makes it difficult to understand why the model predicts physically invalid values about sea ice conditions (e.g., negative SIC or SIC $>$ 100\%). 





In order to address such issues of the fully data-driven ML models and improve the model's generalizability, it is helpful to guide the learning process to agree with fundamental physical laws and domain knowledge. By integrating physics knowledge and constraints into ML training, the ML models can yield physically consistent predictions even in the presence of imperfect data, such as missing, noise, or outliers data \citep{Karniadakis2021, Rassi2019_PINN, Maier2023_mythANN}. This integration of ML and physics knowledge is often referred to as \emph{physics-informed machine learning (PIML)}, and has been proposed for various dynamic systems \citep{Karniadakis2021, Maier2023_mythANN}. Recently, many studies have attempted this PIML framework for various applications including, but not limited to, flow dynamics, material sciences, molecular simulations, chemistry \citep{Karniadakis2021}.
In particular, PIML based on well-known flow laws such as Stokes equation has shown great success in modeling the dynamics of ice sheets and glaciers (i.e., land ice) \citep{Teisberg2021, Riel2021, Riel2023, Iwasaki2023, Jouvet2023, HE2023}. In sea ice application, \cite{LIU2024_PINN} proposed dual-task neural network architecture and incorporated a loss function based on sea ice control equation. However, it remains unclear how this PIML concept and physics-informed loss function can contribute to future sea ice predictions under rapidly changing climate conditions.

This study aims to improve the fidelity of deep learning models that predict sea ice dynamics by integrating physical knowledge into the model training. In order to enforce the deep learning model (neural network) to converge into physically valid SIV and SIC values, we explicitly adopt a physics-informed neural network (PINN) approach based on loss function. We embed this physics-informed learning strategy in the hierarchical information-sharing U-net (HIS-Unet), a CNN model for SIV and SIC predictions \citep{koo2023_hisunet}. The main contributions of this research consist of the following.
\begin{itemize}
\item We design physics loss functions and combine them with the data loss function to regulate physical validity of SIC and SIV.
\item We modify the output layer to guarantee the physically valid SIC values.
\item Our extensive experiments show that the physics-informed deep learning model can improve the performance in SIC and SIV predictions even with a small number of training samples.
\item We explore the spatiotemporal variability in the improved performance of our physics-informed learning strategies compared to fully data-driven deep learning models that do not embed physics.
\end{itemize}

The remainder of the paper is organized as follows. Section \ref{related_work} provides a brief review of the physical background of sea ice modeling, machine learning for sea ice prediction, and PIML. Section \ref{data} explains details of remote sensing and meteorological data used in this study, and section \ref{method} presents the architecture of our CNN model and physics-informed learning strategies. The performance and implication of our PINN are discussed in Section \ref{result}.

\section{Background}\label{related_work}

\subsection{Physics of sea ice dynamics}\label{sea_ice_dynamics}

In physical sea ice models, the volume and area of sea ice are determined by thermodynamic evolution and dynamic motion field. The spatiotemporal changes in SIT ($h$) and SIC ($A$) can be expressed by the following equations for mass conservation \citep{Hibler1979, Holland2012, flato2004}:

\begin{equation}\label{eq_massbalance_SIT}
\frac{\partial h}{\partial t} + \nabla\cdot(\boldsymbol{u}h)=S_h
\end{equation}
\begin{equation}\label{eq_massbalance_SIC}
\frac{\partial A}{\partial t} + \nabla\cdot(\boldsymbol{u}A)=S_A 
\end{equation}
where $\boldsymbol{u}$ is the ice motion vector, $S_h$ and $S_A$ are the changes in SIT and SIC driven by thermodynamic sources (e.g., freezing or melting), respectively.

Additionally, the following momentum equations have been used in various numerous physical sea ice models to explain the balance of horizontal forces on sea ice based on the most common assumption of elastic-viscous-plastic (EVP) properties of sea ice \citep{Hibler1979}:
\begin{equation}\label{eq_momentum}
m\frac{D \boldsymbol{u}}{D t} = -mf\boldsymbol{k}\times\boldsymbol{u} + \tau_{ai} + \tau_{wi} + \boldsymbol{F} -mg \nabla H
\end{equation}
where $D/Dt=\partial/\partial t + \boldsymbol{u} \cdot \nabla$ is the substantial time derivative, $m$ is the ice mass per unit area, $\boldsymbol{k}$ is a unit vector normal to the surface, $\boldsymbol{u}$ is the ice velocity, $f$ is the Coriolis parameter, $\tau_{ai}$ and $\tau_{wi}$ are the forces due to air and water stresses, $H$ is the elevation of the sea surface, $g$ is the gravity acceleration, and $\boldsymbol{F}$ is the force due to variations in internal ice stress. Following this equation, many previous studies described SIV as interacting with wind and ocean forcings \citep{netXtSIM2016, FESOM2014, TIMMERMANN2009, SALASMELIA2002}. Particularly, wind velocity has been treated as a major variable in SIV, contributing to up to 70 \% of the sea ice velocity variances \citep{Thorndike1982} depending on season or region. Nevertheless, predicting sea ice dynamics based on physical models is still challenging due to the intrinsic complexity and dependency of physical models on numerous parameterizations.

\subsection{Neural network for sea ice prediction}\label{NN_for_seaice}

Convolutional neural networks (CNNs) have been used as the most popular and efficient deep learning network for modeling SIC and SIV. First, in terms of SIC, \cite{Andersson2021} proposed a deep-learning sea ice forecasting system named IceNet to forecast monthly SIC for the next six months. \cite{Grigoryev2022} used a U-Net architecture, a type of deep CNN, for daily SIC forecasts in several subsections of the Arctic Ocean, including the Barents and Kara Seas, Labrador Sea, and Laptev Sea. \cite{Kim2020} also used CNN to predict monthly SIC from satellite-based SIC observations, sea surface temperature, air temperature, albedo, and wind velocity data from the previous months. Similarly, \cite{Fritzner2020} proposed a CNN that predicts weekly SIC from SIC, sea surface temperature, and air temperature from the previous six days. CNN models proposed by \cite{Ren2021} and \cite{Ren2022} used the sequences of satellite-derived SIC observations to make daily SIC predictions. In some studies, long short-term memory (LSTM), an advanced recurrent neural network (RNN), has been modified and inserted into CNN architecture to obtain better performance in both spatial and temporal SIC predictions \citep{Liu2021_daily}. CNN and LSTM have also been used for SIV prediction in several studies. \cite{zhai2021} and \cite{Hoffman2023} used CNN to predict daily SIV, and their model outperformed other statistical and physical models. \cite{Petrou2019} showed that adding LSTM units to convolutional layers improves the performance of a model that predicts SIV. In this study, we adopt a multi-task neural network framework named HIS-Unet \citep{koo2023_hisunet} and improve the predictability of SIV and SIC of this model by incorporating physics-informed training scheme.

\subsection{Physics-informed machine learning (PIML)}

Historically, PIML has been achieved by integrating physics knowledge into ML training in the form of: (i) data, (ii) model architecture, and (iii) optimization \citep{Hao2023_PIML, Karniadakis2021}. First, one of the straightforward methods to incorporate physics knowledge into ML is to generate training data from the desired physics knowledge. Data-driven ML models that are trained with sufficient simulation data governed by certain underlying physics laws can accurately represent these physics laws. Given that enough high-quality and labeled data is not always available for real-world tasks, synthetic data constructed with physical simulations can provide a large amount of data with high quality. However, this approach can have limitations in perfectly reflecting real world because real-world data have different distributions from simulation data. In addition, the cost of data acquisition can be a critical issue, as simulation data is often generated via expensive experiments or large-scale simulations.


The second way to implement PIML is to design neural network architectures that implicitly embed any prior knowledge and inductive biases associated with a given task (e.g., symmetry, conservation laws, etc.) \citep{Karniadakis2021}. CNN is a popular example of this neural network architecture that achieves extensive applicability for image recognition by respecting invariance and symmetry groups in natural images \citep{Karniadakis2021, Mallat2015}. Another example of this category is equivariant networks, which embed the dynamic changes in spatial coordinates to preserve the equivariance of data points to rotation and translation \citep{Satorras2022_egcn, Schutt_2017_Schnet}. Some neural network architectures also used Lagrangian and Hamiltonian mechanics to enforce the energy conservation property of the networks \citep{Hao2023_PIML}. Furthermore, in solving PDEs, there have been several attempts to modify neural network architecture to satisfy the required initial conditions \citep{Hao2023_PIML}. Although such a model architecture approach can be effective with relatively simple and well-defined physics or symmetry groups, this approach has limitations in extending to highly complex problems. 

Lastly, physics knowledge can be integrated into the optimization and learning process by imposing constraints of prior physics knowledge into the form of loss functions. As a soft manner of penalizing the loss function instead of enforcing a specific condition directly, this approach can indirectly reshape the target spaces of NN output to converge to physically plausible solutions \citep{Karniadakis2021, Hao2023_PIML}; in this study, we will refer to this approach specifically as \emph{PINN}. This approach can be regarded as a case of multi-task learning, balancing the loss functions for two tasks of fitting both the observation data and physical constraints. Although such a soft constraint approach is the most common and flexible way for PIML, balancing these two tasks can be challenging because they can counteract the convergence to each other's solution.

In the modeling of land ice, PIML has emerged as an effective way to predict ice flow, and most previous studies employed the optimization approach by adding physics loss functions or regularization terms (i.e., PINN). \cite{Teisberg2021} developed a PINN to predict ice thickness and velocity by adding a physics loss function based on the mass conservation of ice sheets. \cite{Riel2021} introduced a physics loss function derived from the governing equations of ice flow to their PINN framework to infer spatially and temporally varying basal drag in ice sheets. \cite{Riel2023} proposed a PINN framework to predict the distribution of ice rigidity by adding the loss functions of ice flow and Kullback-Leibler divergence. \cite{Iwasaki2023} proposed a physics loss function to fit the model results with physics laws regarding ice flows, and this physics loss function contributed to improving the model performance when the data was contaminated by noise. \cite{Jouvet2023} optimized their PINN by minimizing the energy associated with high-order ice-flow equations. \cite{Cheng2024} employed a PINN to infer basal sliding while filling gaps in sparsely observed ice thickness data. In the regime of sea ice, there have been several attempts to integrate physics knowledge into machine learning by training the model with the data retrieved from physical models \citep{Palerme2021, Palerme2024}. \cite{LIU2024_PINN} used a dual-task CNN architecture and physics-informed loss function to enforce dynamic constraints of SIC and SIV. Despite such developments of PINN for cryosphere, it is still necessary to assess how PINN and physical constraints can help future predictions of sea ice conditions. Therefore, this study explicitly applies the concept of PINN to sea ice prediction and explore the how weights to physics-informed loss function and representability of training samples contribute to the model predictability for future unseen sea ice conditions.



\section{Data}\label{data}

In this study, we use daily SIV and SIC data collected from satellite observation from 2009 to 2022. We use the SIV and SIC from previous three days (inputs of the model) to predict the next day's SIV and SIC (output of the model) (Table \ref{table1}). Besides satellite observations of SIV and SIC, we also use wind velocity and air temperature from reanalysis as additional input variables for the model. Additionally, we add X and Y coordinates as inputs to represent regional variability. The input datasets and their sources are summarized in Table \ref{table1}. This section presents the details of these datasets. 

\begin{table*}
\centering
\caption{Input data sets for machine learning models}
\begin{tabular}{m{3cm}m{8cm}m{2cm}}
\hline
Dataset & Name& \makecell{Spatial\\resolution} \\ 
\hline
Sea ice velocity (u and v)& NSIDC Polar Pathfinder Daily 25 km EASE-Grid Sea Ice Motion Vectors \citep{Tschudi2020} & 25 km \\ 
Sea Ice Concentration & NOAA/NSIDC Climate Data Record of Passive Microwave Sea Ice Concentration \citep{Meier2021} & 25 km \\ 
10 m wind velocity (u and v) & ECMWF Reanalysis v5 (ERA5) hourly data on single levels \citep{ERA5} & 0.25$^{\circ}$ \\ 
2 m air temperature & ECMWF Reanalysis v5 (ERA5) hourly data on single levels \citep{ERA5} & 0.25$^{\circ}$ \\ 
\hline
\end{tabular}
\label{table1}
\end{table*}

\subsection{Sea ice velocity}
We use the NSIDC Polar Pathfinder Daily 25 km EASE-Grid Sea Ice Motion Vectors version 4 \citep{NSIDC, Tschudi2020} as the input and output SIV. This product derives daily SIV from three primary types of sources: (1) gridded satellite images, (2) wind reanalysis data, and (3) buoy position data from the International Arctic Buoy Program \citep{NSIDC, Tschudi2020}. The u component (along-x) and v component (along-y) of SIV are independently derived from each of these sources and optimally interpolated onto a 25 km Equal-Area Scalable Earth (EASE) grid. When SIV is derived from satellite images, a correlation coefficient is calculated between a small target area in a one-day image and a searching area in the next-day image. Then, the location in the next-day image with the highest correlation coefficient is determined as the displacement of ice \citep{NSIDC}. The mean difference between the Polar Pathfinder and buoy measurements of SIV is approximately 0.1 km/day and 0.3 km/day for u and v components, respectively \citep{NSIDC}. We note that the SIV of this product is valid over short distances away from the ice edge in areas where ice conditions are relatively stable, stationary, homogenous, and isotropic \citep{Tschudi2020}. In this study, we exclude SIV values close to the coastlines within 50 km (or 2 pixels) of distance.



\subsection{Sea ice concentration}
For the SIC data, we use NOAA/NSIDC Climate Data Record of Passive Microwave Sea Ice Concentration version 4 data \citep{Meier2021}. This data set provides a Climate Data Record (CDR) of SIC (i.e., the areal fraction of ice within a grid cell) from passive microwave (PMW) data. The CDR algorithm output is the combination of SIC estimations from two algorithms: the NASA Team algorithm \citep{NASA_team} and NASA Bootstrap algorithm \citep{NASA_BT}. These empirical algorithms estimate SIC from the PMW brightness temperatures at different frequencies and polarizations: vertical and horizontal polarizations at 19 GHz, 22 GHz, and 37 GHz. Several assessments showed that the error of this SIC estimation is approximately 5 \% within the consolidated ice pack during cold winter conditions \citep{Meier2005, Comiso1997, Ivanova2015}. However, in the summer season, the error can rise to more than 20 \%  due to surface melt and the presence of melt ponds \citep{Kern2020}. Due to the data quality issue near coastal areas, we use the SIC data more than 50 km from the coastline. To match the coordinate systems of the SIC and SIV products, we reproject the NSIDC Sea Ice Polar Stereographic grid of the SIC product into the EASE grid of the SIV product using bilinear interpolation.


\subsection{ERA5 climate reanalysis}
As shown in Eq. \ref{eq_momentum}, sea ice dynamics are largely associated with atmospheric and oceanic circulation. Thus, we use the wind speed and air temperature as the input variables of the ML model. We use the fifth-generation ECMWF (European Centre for Medium-Range Weather Forecasts) atmospheric reanalysis (ERA5) as the data sources for wind velocity and air temperature. ERA5 provides hourly estimates of atmospheric, land, and oceanic climate variables, covering the period from January 1940 to the present \citep{ERA5}. We obtain the daily average wind velocity (zonal and meridional components) at 10 m height and 2 m air temperature from this hourly data. To co-locate this data with the SIV and SIC data, we reproject the raw ERA5 latitude-longitude grid onto the 25 km EASE grid using bilinear interpolation.

\section{Methods}\label{method}

In this study, we use a hierarchical information-sharing U-net (HIS-Unet) based on \cite{koo2023_hisunet} as a backbone deep learning model (Fig. \ref{HISUnet}) to make daily predictions of SIC and SIV from inputs of SIC, SIV, wind, and atmospheric temperature from the previous three days. As a fully convolutional network, the HIS-Unet is designed to predict SIC and SIV simultaneously. By sharing SIC and SIV information during the propagation processes, the HIS-Unet achieves better fidelity than other neural networks and statistical approaches in predicting both SIC and SIV. In particular, this information sharing is proven to be useful for predicting sea ice conditions where and when SIV has impacts on SIC changes \citep{koo2023_hisunet}. Since understanding sea ice dynamics requires both SIV and SIC information (e.g., Eq. \ref{eq_massbalance_SIC}), we use the multi-task prediction of SIV and SIC by the HIS-Unet to facilitate integrating the physics knowledge of sea ice dynamics. To embed fundamental knowledge of sea ice dynamics into the HIS-Unet architecture, we (1) introduce physics loss functions in the training phase and (2) add an additional activation function to the output layer. This section presents the details of the HIS-Unet architecture and physics-informed training strategies used in this study.

\subsection{Hierarchical information-sharing U-net (HIS-Unet)}

\begin{figure}
\centerline{\includegraphics[width=1.0\columnwidth]{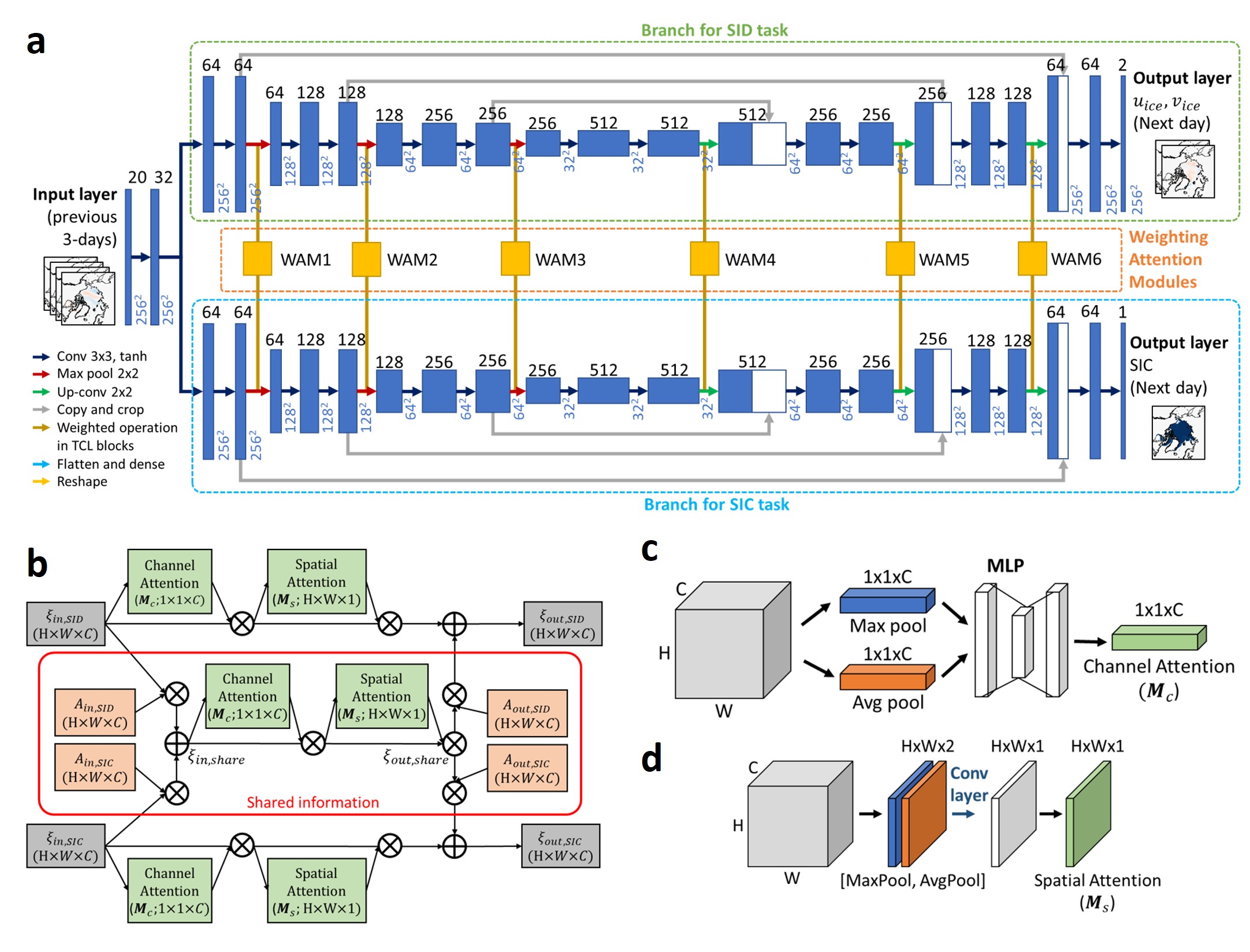}}
\caption{(a) Architecture of Hierarchical information-sharing U-net (HIS-Unet); (b) Weighting attention module (WAM) in HIS-Unet; (c) Channel attention module and (d) Spatial attention module in a WAM \citep{koo2023_hisunet}.}
\label{HISUnet}
\end{figure}

The HIS-Unet architecture consists of two separate task branches, each for predicting SIV and SIC (Fig. \ref{HISUnet}). The task branches are connected with weighting attention modules (WAMS) that allow information sharing between the SIV and SIC branches. SIC and SIV branches have U-net structures \citep{Ronneberger2015} with a series of encoders (contracting path) and decoders (expansive path). Each encoder path consists of repeated two $3\times3$ convolutions and a $2\times2$ max pooling operation with stride 2 for downsampling and doubling the number of features. The decoders conduct $2\times2$ up-convolution and half the number of feature channels, followed by a concatenation with the cropped feature map and two $3\times3$ convolutions \citep{Ronneberger2015}. The hyperbolic tangent (tanh) activation function is applied after each convolutional layer:
\begin{equation}\label{eq:tanh}
Tanh(x) = \frac{e^x-e^{-x}}{e^x+e^{-x}}
\end{equation}

The separated U-net structures of SIV and SIC branches output SIV or SIC, respectively, but they share and transfer their information through six weighting attention modules (WAMs) during the propagation process. Six WAMs are inserted into 3 encoder steps and 3 decoder steps between SIC and SIV branches (Fig. \ref{HISUnet}). 

Each WAM first receives information from the SIV and SIC branches and calculates the weighted sum of them (Fig. \ref{HISUnet}b). Letting a WAM receive the SIV feature map ($\xi_{in,SIV}$; height $H$, width $W$, channels $C$) and SIC feature map ($\xi_{in,SIC}$; $H \times W \times C$), the input shared information ($\xi_{in,share}$) is determined by multiplying linear weights, $A_{in,SIV}$ and $A_{in,SIC}$, to $\xi_{in,SIV}$ and $\xi_{in,SIC}$, respectively (Fig. \ref{HISUnet}b). Then, this shared information $\xi_{in,share}$ passes through sequentially arranged channel attention (Fig. \ref{HISUnet}c) and spatial attention modules (Fig. \ref{HISUnet}d). The channel attention highlights what channel is meaningful, and the spatial attention highlights where an informative part is spatially located \citep{Woo2018}. These channel and spatial attention modules are also applied to the input SIV and SIC feature maps ($\xi_{in,SIV}$ and $\xi_{in,SIC}$) (Fig. \ref{HISUnet}b). Then, the attention shared information is sent to the SIV and SIC branches after multiplying output weights ($A_{out,SIV}$ and $A_{out,SIC}$) and adding attention SIV and SIC information, respectively. More details of how the WAMs enable sharing and highlighting SIC and SIV information are described in \cite{koo2023_hisunet}.

\subsection{Physics-informed training}

In order to make the HIS-Unet incorporate physics knowledge of sea ice, we embed two physics-informed regularizations: (1) include physics loss functions along with the data loss function and (2) insert the sigmoid activation function to the SIC branch to guarantee valid SIC values.

The original HIS-Unet is optimized by the mean square error (MSE) objective loss function ($L_{data}$). The MSE data loss term ($L_{data}$) is calculated by the following equation:

\begin{equation}\label{loss_data}
L_{data} = \sum (\lvert u_{p}-u_{o} \rvert^2 + \lvert v_{p}-v_{o} \rvert^2 + \lvert A_{p}-A_{o} \rvert^2)
\end{equation}
where $u$ and $v$ denote x-component and y-component of SIV, respectively, $A$ denotes SIC, and the subscript $o$ means observation and $p$ means prediction by HIS-Unet. In addition to this MSE data loss function, we design physics loss functions that apply physical constraints: (i) $L_{sat}$ constraining valid SIV values and (ii) $L_{therm}$ constraining thermodynamically valid ice growth. 

First, we regularize the valid SIV values associated with SIC values. Since the PMW-derived SIV can be only defined where sea ice presents with $>$ 15 \% of SIC \citep{NSIDC}, SIV should be zero where SIC is less than 15 \%. Therefore, we design the first physics loss term as follows:

\begin{equation}\label{loss_phy1}
L_{sat}=
\begin{cases}
    |u_{p}^2+v_{p}^2|,& \text{if } A_p < 0.15\\
    0, & \text{if } A_p \geq 0.15
\end{cases}
\end{equation}

The next physics loss term ($L_{therm}$) is based on Eq.\ref{eq_massbalance_SIC}, which explains the thermodynamic and dynamic SIC changes. The thermodynamic SIC changes ($S_A$), the right term in Eq. \ref{eq_massbalance_SIC}, can be calculated by temporal changes of SIC and the combination of advection and divergence of SIC. In the case of daily SIC prediction, the daily $S_A$ can be assumed not to exceed (-1, 1). That is, we assume that the thermodynamic freezing and melting of sea ice are unlikely to saturate SIC from 0 \% to 100 \% or remove the entire sea ice from 100 \% to 0 \% within a day in most cases. Thus, based on this assumption, we design the second physics loss term as follows:
\begin{equation}\label{loss_phy2}
L_{therm}=\text{ReLU}(|\frac{\partial A_p}{\partial t} + \nabla\cdot(\boldsymbol{u}_pA_p)|-1)
\end{equation}
where the time derivative term ($\frac {\partial A_p}{\partial t}$) is derived by subtracting the previous-day SIC from the output SIC, and the spatial derivative term ($\nabla\cdot(\boldsymbol{u}_pA_p)$) is derived by calculating the spatial gradients in output SIV and SIC grids. By using $L_{therm}$, we can assign a linearly increasing penalty as the thermodynamic ice growth prediction exceeds 1.

Consequently, the total physics loss term ($L_{phy}$) and the final objective loss functions ($L$) are defined as follows:
\begin{equation}\label{loss_phy}
L_{phy}=\lambda_{sat}L_{sat}+\lambda_{therm}L_{therm}
\end{equation}
\begin{equation}\label{loss_total}
L=L_{data}+L_{phy}
\end{equation}
where $\lambda_{sat}$ and $\lambda_{therm}$ are the relative weights of $L_{sat}$ and $L_{therm}$, respectively, to the data loss term. In this study, we conduct experiments with different $\lambda_{sat}$ and $\lambda_{therm}$ of 0, 0.2, 1.0, and 5.0 and examine how this weight changes the model performance.

As the second regularization, we enforce the output SIC to lie within the value range of 0 to 1 (0 to 100 \%) by adding the sigmoid activation function to the last output layer of the SIC branch:
\begin{equation}\label{eq:sigmoid}
Sigmoid(x) = \frac{1}{1+e^{-x}}
\end{equation}
This can constrain the valid range of SIC outputs, even for out-of-training samples. 

\subsection{Traininig strategy}

In our HIS-Unet, we use the previous 3 days of SIV (x- and y-components), SIC, air temperature, and wind velocity (x- and y- components) as the inputs to predict the next day's SIV and SIC. Consequently, the input layer has 18 channels of 256$\times$256 grid size. All input values are normalized to -1 to 1 based on the nominal maximum and minimum values that each variable can have. All the data is collected for 14 years from 2009 to 2022; the first seven years of data (2009-2015) are utilized as training data, and the remaining seven years of data (2016-2022) are utilized as test data. To examine how the model performance is changed by the number of training data, we train the model with three different training sample sizes: (1) using all 2009-2015 data as training samples (i.e., 100 \% sampling), (2) randomly selecting 50 \% of 2009-2015 data as training samples (i.e., 50 \% sampling), and (3) randomly selecting 20 \% of 2009-2015 data as training samples (i.e., 20 \% sampling). The total number of training samples is 2,177 for the 100 \% sampling, 1,088 for 50 \% sampling, and 436 for the 20 \% sampling. The HIS-Unet models trained with three different sample sizes are applied to the 2016-2022 test data. All models are optimized by the Adam stochastic gradient descent algorithm \citep{Kingma2015} with 100 epochs and a 0.001 learning rate. All scripts are executed on eight NVIDIA RTX A5000 GPUs with 24 GB of memory.

\subsection{Model performance}

The model performance is assessed by the root mean square error (RMSE):
\begin{equation}
\text{RMSE}(\hat{y}, y)=\sqrt{\frac{\sum_{i=1}^N(\hat{y}_i-y_i)^2}{N}}
\label{RMSE}
\end{equation}
where $\hat{y}$ denotes predicted values, $y$ denotes true values, $N$ is the number of data points. In the case of SIV, we calculate RMSE for both x- and y-component SIV, and the average of x- and y-component RMSEs is determined as the SIV RMSE to embrace the magnitude and angle error of SIV. We assess and compare the RMSEs from the HIS-Unet with three different training sample cases and four different $\lambda_{sat}$ and $\lambda_{therm}$ (i.e., 0, 0.2, 1.0, and 5.0). We use the HIS-Unet without any physics-informed regularization (i.e., trained only with the data loss $L_{data}$ and without the sigmoid activation function to the SIC branch) as the baseline model, and this purely data-driven model is notated as \emph{No-Phy}. Meanwhile, we call the physics-informed HIS-Unet simply PINN. We compare the RMSE differences between PINN and \emph{No-Phy} for different training sample sizes (20 \%, 50 \%, and 100 \%). Additionally, we examine how these RMSE differences vary by month and region in seven test years (2019-2022). To assess the statistical significance of RMSE improvement by PINN, we conduct paired t-tests between RMSEs of PINN and RMSEs of \emph{No-Phy}. If the p-value from the t-test is below 0.05, we determine that the RMSE difference between PINN and \emph{No-Phy} is significant.

In addition to RMSE, we use the mean absolute error (MAE) and anomaly correlation coefficient (ACC) to evaluate the predictive performance of PINN and \emph{No-Phy}.
\begin{equation}
\text{MAE}(\hat{y}, y)=\frac{\sum_{i=1}^N |\hat{y}_i-y_i|}{N}
\label{MAE}
\end{equation}
\begin{equation}
\text{ACC}(\hat{y}, y)=\frac{\sum_{i=1}^N(\hat{y}_i-\bar{\hat{y}_i})(y_i-\bar{y_i})}{\sqrt{\sum_{i=1}^N(\hat{y}_i-\bar{\hat{y}_i})^2\sum_{i=1}^N(y_i-\bar{y_i})^2}}
\label{ACC}
\end{equation}
Lower RMSE and MAE values indicate higher prediction accuracy, while a higher ACC (i.e., approaching 1) reflects stronger agreement between predicted and observed anomalies, indicating better predictive skill. We note that the performance of \emph{No-Phy} model trained with full training samples can be found in \cite{koo2023_hisunet} in detail. The \emph{No-Phy} HIS-Unet showed better performance than another simple statistical model (e.g., linear regression), physical model (Hybrid Coordinate Ocean Model; HyCOM), simple convolutional network, and independent-branch U-nets. Since the HIS-Unet performance has already been assessed in the previous study, this study is focused on assessing the effect of using a physics-informed learning strategy rather than assessing the performance of the HIS-Unet architecture itself.

\section{Results and discussion}\label{result}

\subsection{Performance of PINN}

\begin{figure}
\centerline{\includegraphics[width=1.0\columnwidth]{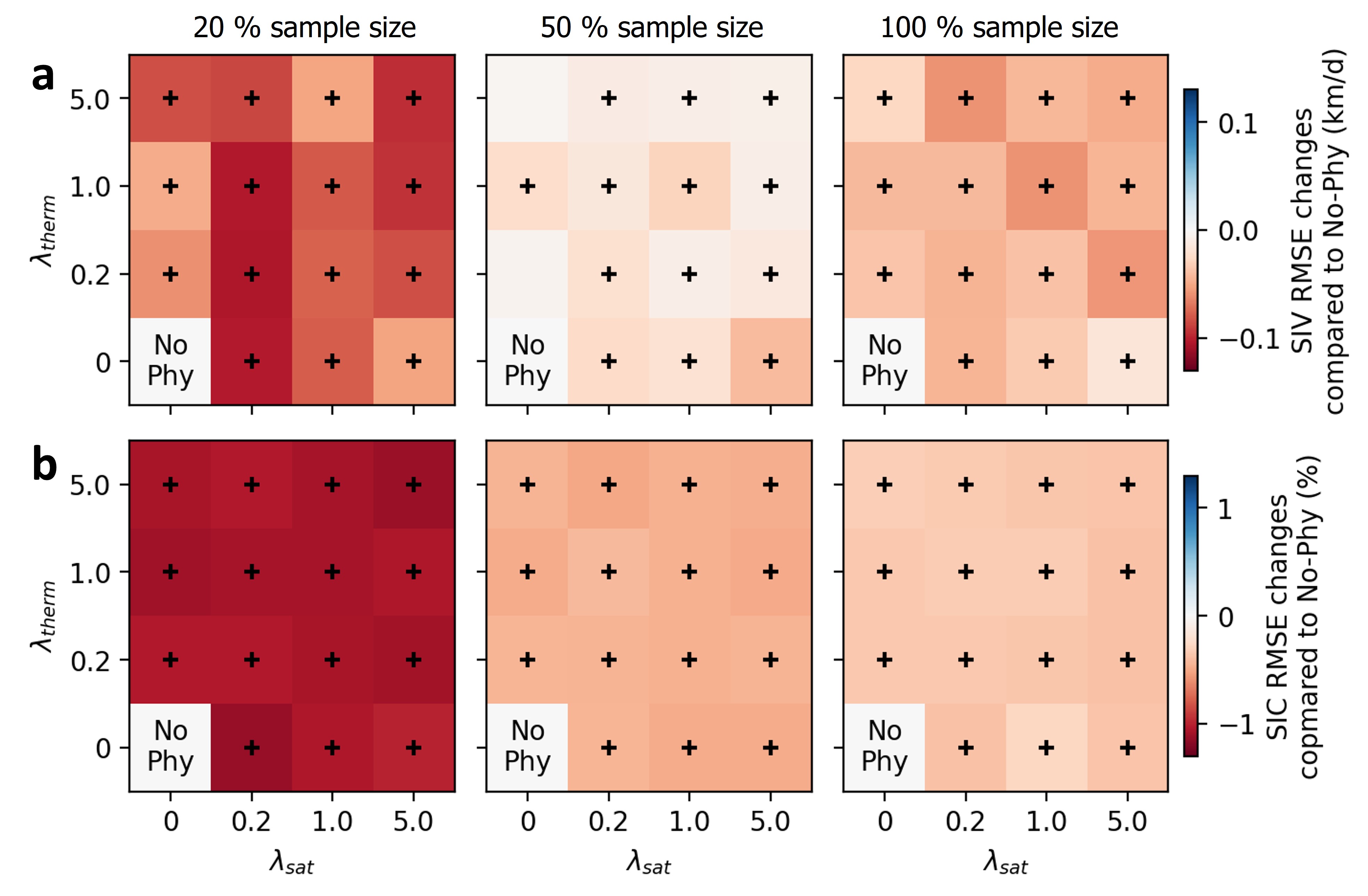}}
\caption{(a) The reduction of SIV RMSE by PINNs with different training sample sizes (20, 50, and 100 \%) and combinations of $\lambda_{sat}$ and $\lambda_{therm}$ (0, 0.2, 1.0, and 5.0), relative to the SIV RMSE of the \emph{No-Phy} model. (b) The reduction of SIC RMSE by PINNs with  different training sample sizes and combinations of $\lambda_{sat}$ and $\lambda_{therm}$, relative to the SIC RMSE of the \emph{No-Phy} model. Cross markers indicate statistically significant reduction.}
\label{rmse_total}
\end{figure}

We evaluate the predictive performance of \emph{No-Phy} and PINNs with different $\lambda_{sat}$ and $\lambda_{therm}$ settings, and check whether PINNs can improve the model performance compared to \emph{No-Phy}. Additionally, we investigate how this improvement pattern varies by different training sample sizes. Figure \ref{rmse_total} shows the SIV and SIC RMSE changes by PINNs as functions of different training sample sizes, $\lambda_{sat}$, and $\lambda_{therm}$. Table \ref{table_accuracy} shows the RMSE, MAE, and ACC of \emph{No-Phy} baseline and PINN with 0.2 $\lambda_{therm}$ and 0.2 $\lambda_{sat}$. According to Table \ref{table_accuracy}, the errors of both SIV and SIC predictions decrease with more training samples.

Regarding SIV, the integration of physics loss functions, including both $\lambda_{therm}$ and $\lambda_{sat}$, improves the model fidelity relative to the \emph{No-Phy} baseline model (Fig. \ref{rmse_total}a). The reduction in SIV RMSE by the physics loss term is observed in all sample sizes, but this effect is more significant in the 20 \% sample size: SIV RMSE decreases by up to 0.10 km/day when $\lambda_{sat}$ is set to 0.2. When 50 \% and 100 \% of training samples are used, the SIV RMSE decreases by up to 0.02 km/day and 0.05 km/day, respectively. We also find improvements of SIV MAE and ACC by PINN in all training sample cases (Table \ref{table_accuracy}).


On the other hand, regarding SIC, the PINNs show lower SIC RMSEs than the \emph{No-Phy} baseline model (Fig. \ref{rmse_total}), and the improvement of the PINN is more significant for smaller sample sizes. When 20 \% of training samples are used, the PINNs show approximately 1.1 \% lower SIC RMSE than the \emph{No-Phy} model. When the training samples are set to 50 \% and 100 \%, the SIC RMSEs are approximately 0.5 \% and 0.3 \% lower for the PINNs than the \emph{No-Phy} models, respectively. Besides the RMSE comparison, the MAE and ACC are also improved by PINN (Table \ref{table_accuracy}). These results suggest that the incorporation of physics-informed architecture and optimization is beneficial for SIC predictability under limited training sample scenarios. However, the weights to the physics loss function are not so significant for the SIC accuracy: no consistent trend in SIC RMSE is observed with increasing $\lambda_{therm}$ or $\lambda_{sat}$ values.

\begin{table*}\scriptsize
\centering
\caption{Assessment of SIV and SIC predictions for PINN ($\lambda_{therm}=0.2$ and $\lambda_{sat}=0.2$) and \emph{No-Phy} with different training sample sizes. The best accuracy is highlighted in bold.}
\begin{tabular}[t]{c|c|ccc|ccc|ccc}
\hline
\multicolumn{2}{c|}{Sample size} & \multicolumn{3}{c|}{20 \%} & \multicolumn{3}{c|}{50 \%} & \multicolumn{3}{c}{100 \%} \\
\cline{1-11}
\multirow{3}{*}{SIV} & Model & \makecell{RMSE\\(km/d)} & \makecell{MAE\\(km/d)} &ACC & \makecell{RMSE\\(km/d)} & \makecell{MAE\\(km/d)} &ACC & \makecell{RMSE\\(km/d)} & \makecell{MAE\\(km/d)} &ACC \\
\cline{2-11}
& PINN   & 2.778 & 1.954 & 0.853 & 2.732 & 1.915 & 0.859 & \textbf{2.629} & \textbf{1.844} & \textbf{0.867} \\
& No-Phy & 2.873 & 2.053 & 0.843 & 2.759 & 1.936 & 0.856 & 2.684 & 1.880 & 0.863 \\

\hline
\multirow{3}{*}{SIC} & Model & \makecell{RMSE\\(\%)} & \makecell{MAE\\(\%)} &ACC & \makecell{RMSE\\(\%)} & \makecell{MAE\\(\%)} &ACC & \makecell{RMSE\\(\%)} & \makecell{MAE\\(\%)} &ACC \\
\cline{2-11}
& PINN   & 6.687 & 3.441 & 0.975 & 6.222 & 3.118 & 0.978 & \textbf{5.854} & \textbf{2.917} & \textbf{0.981}  \\
& No-Phy & 7.393 & 4.335 & 0.969 & 6.611 & 3.533 & 0.975 & 6.197 & 3.274 & 0.978 \\
\hline
\end{tabular}
\label{table_accuracy}
\end{table*}%

\subsection{Temporal characteristics of model performance}

\begin{figure}
\centerline{\includegraphics[width=1.0\columnwidth]{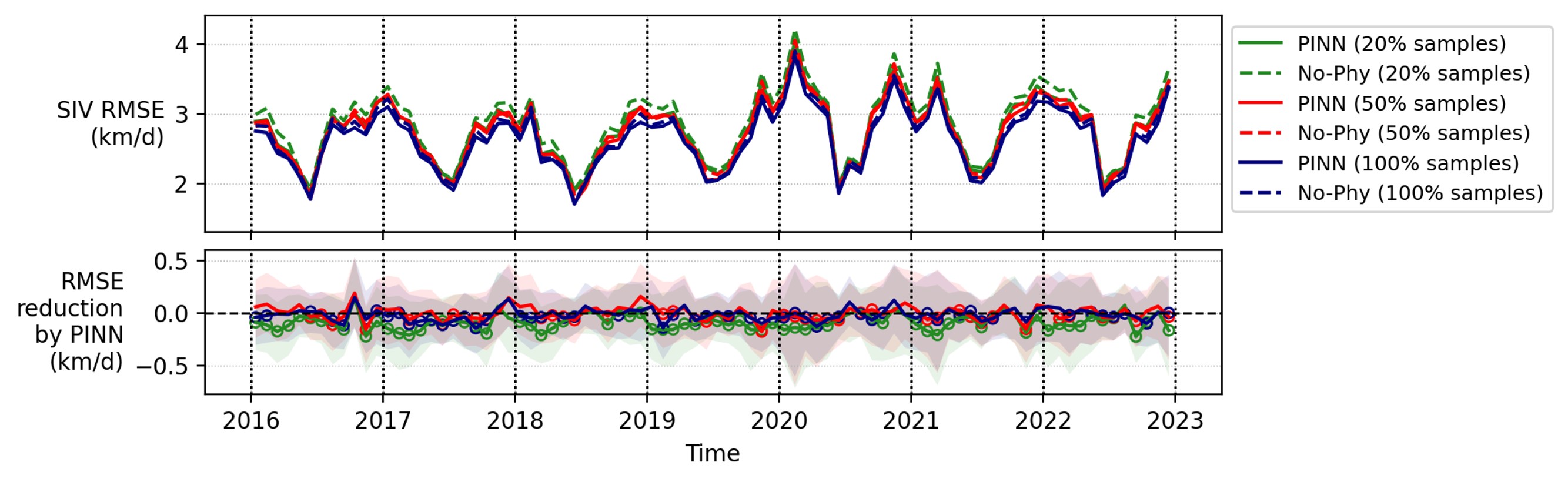}}
\caption{Monthly SIV RMSE of \emph{No-Phy} model and PINN ($\lambda_{sat}=0.2$ and $\lambda_{therm}=0.2$) for seven test years (2016-2022). On the top panel, the solid lines indicate monthly SIV RMSEs of PINN models, and the dashed lines indicate \emph{No-Phy} models. The bottom panel shows the differences between each PINN model and the \emph{No-Phy} model. Statistically significant decreases in RMSE by PINN are highlighted as circles in the bottom panel. Statistical significance of each month is determined by conducting a t-test on all daily RMSE results from PINN and \emph{No-Phy}. The shaded areas in the bottom panels show the 25-75 \% quantile of RMSE difference.}
\label{rmse_siv_month}
\end{figure}

Figure \ref{rmse_siv_month} shows the monthly RMSE of SIV prediction for seven test years (2016-2022) with different sample sizes. For the PINN model, we set $\lambda_{sat}=0.2$ and $\lambda_{therm}=0.2$ based on the optimal configuration identified in Figure \ref{rmse_total}. In general, SIV prediction shows relatively lower errors in summer (June-August) due to the slow drift speed during these months (Fig. \ref{mean_SIC_SIV}c). The improvement of SIV RMSE by PINN (i.e., RMSE difference between PINN and \emph{No-Phy} model) is evident in most months when the sample size is 20 \%, and the RMSE reduction is relatively greater during the winter months (January to April) (Fig. \ref{rmse_siv_month}). For the 50 \% and 100 \% sample sizes, no significant temporal trend in the SIV performance improvement is found.



\begin{figure}
\centerline{\includegraphics[width=1.0\columnwidth]{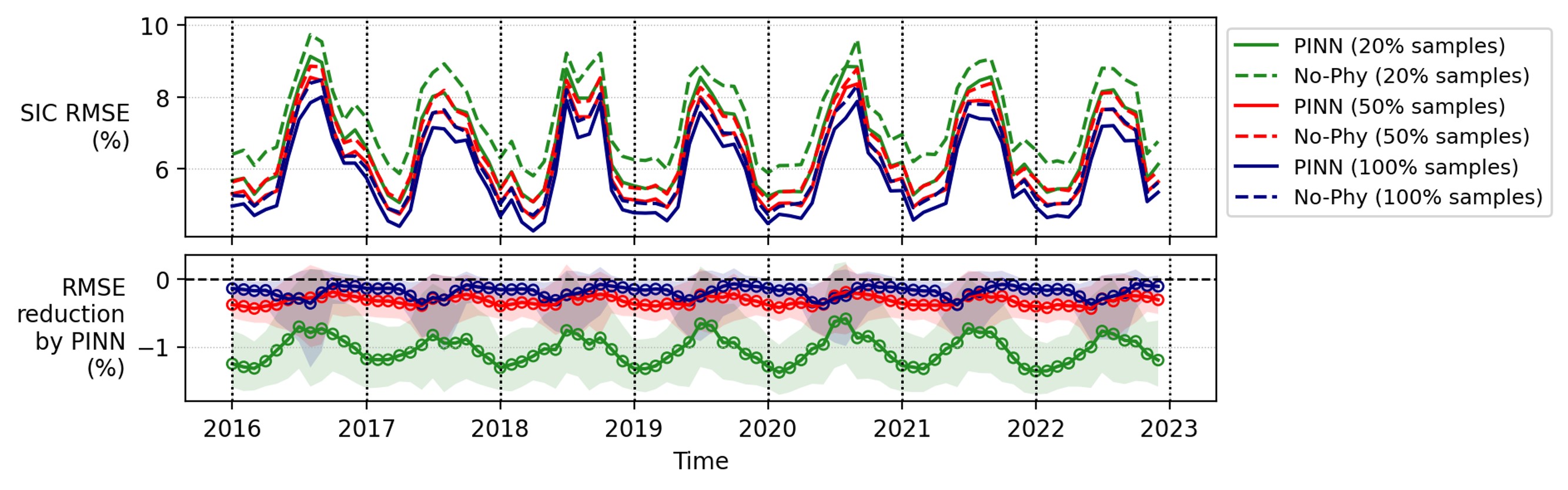}}
\caption{Monthly SIC RMSE of \emph{No-Phy} model and PINN ($\lambda_{sat}=0.2$ and $\lambda_{therm}=0.2$) for seven test years (2016-2022). On the top panel, the solid lines indicate monthly SIC RMSEs of PINN models, and the dashed lines indicate \emph{No-Phy} models. The bottom panel shows the differences between each PINN model and the \emph{No-Phy} model. Statistically significant decreases in RMSE by PINN are highlighted as circles in the bottom panel. Statistical significance of each month is determined by conducting a t-test on all daily RMSE results from PINN and \emph{No-Phy}. The shaded areas in the bottom panels show the 25-75 \% quantile of RMSE difference.}
\label{rmse_sic_month}
\end{figure}

Monthly RMSEs of SIC for test years are depicted in Figure \ref{rmse_sic_month}. In contrast to the SIV results, SIC exhibits relatively higher errors during the summer months from June to July. While SIC remains consistently high from January to May, it decreases rapidly in these months (Fig. \ref{mean_SIC_SIV}d); the model could have difficulty capturing this trend. The most interesting finding is that SIC RMSE is highly dependent on training sample size: SIC RMSE decreases with a greater sample size. As a result, a more remarkable improvement in model accuracy by PINN is observed with fewer sample sizes. When the models are trained with 20 \% of training samples, PINN reduces the SIC RMSE by $>$ 0.5 \% in all months, and such RMSE reduction is more significant in winter months from January to March.


\subsection{Spatial characteristics of model performance}

\begin{figure}
\centerline{\includegraphics[width=1.0\columnwidth]{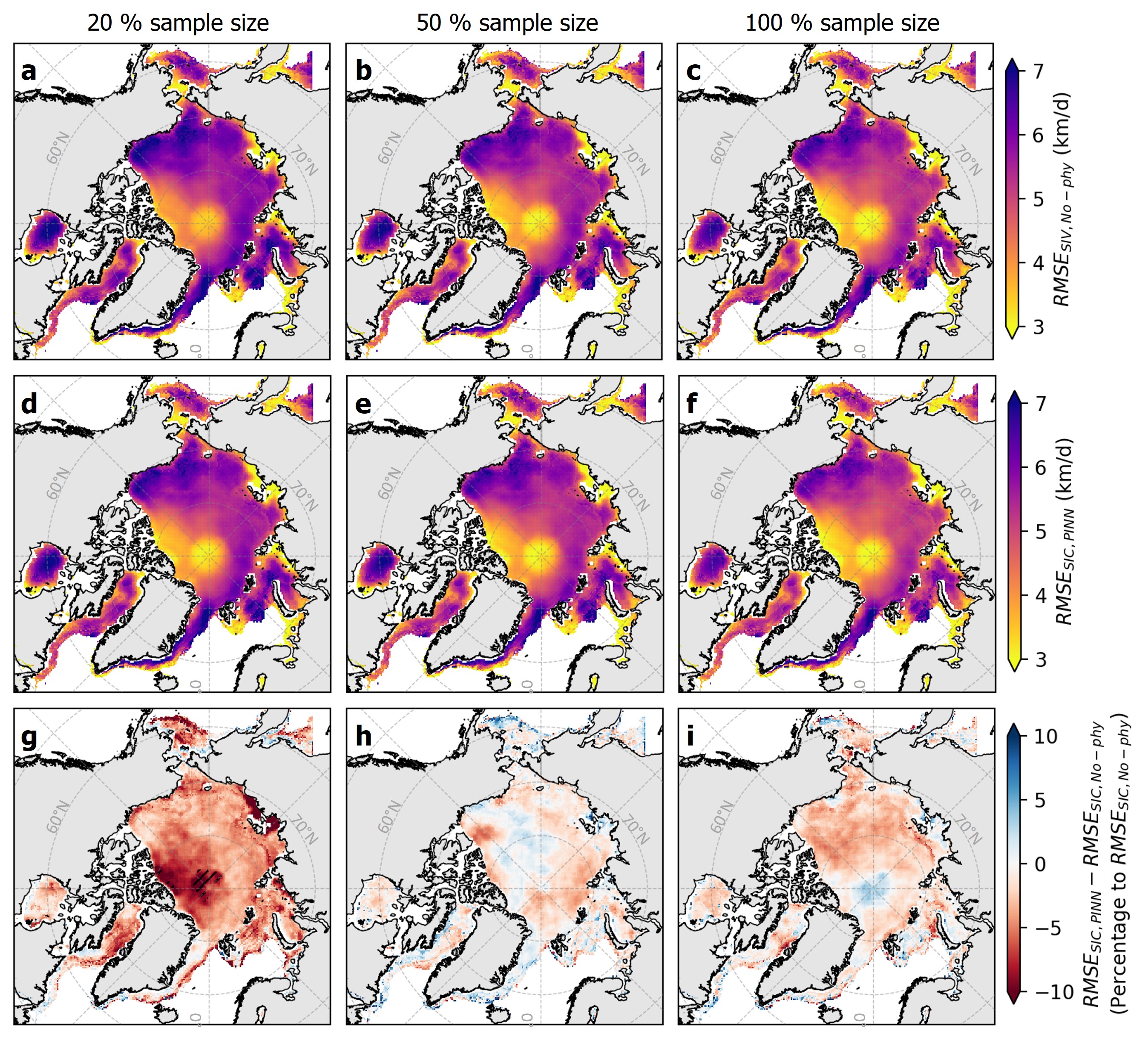}}
\caption{The SIV RMSE map of the No-Phy model for (a) 20 \% training samples, (b) 50 \% training samples, and (c) 100 \% training samples. The SIV RMSE map of the PINN model ($\lambda_{sat}=0.2$ and $\lambda_{therm}=0.2$) for (d) 20 \% training samples, (e) 50 \% training samples, and (f) 100 \% training samples. The relative SIV RMSE difference map between PINN and No-Phy model for (g) 20 \% training samples, (h) 50 \% training samples, and (i) 100 \% training samples. On the bottom panels of the RMSE differences, the reddish color indicates the improvement in SIV RMSE by PINN over the No-Phy model (i.e., reduction in RMSE).}
\label{rmse_siv_map}
\end{figure}

In this section, we examine where the PINN approach is more or less effective and discuss the implications of such spatial patterns. Herein, we discuss only the PINN with $\lambda_{sat}=0.2$ and $\lambda_{therm}=0.2$ configuration because (1) this PINN setting shows overall best RMSE in all sampling sizes and (2) the spatial patterns of PINN improvement appear similar for all different $\lambda_{sat}$ and $\lambda_{therm}$ values. The spatial distributions of SIV RMSE for PINN and \emph{No-Phy} models with different sample sizes are shown in Figure \ref{rmse_siv_map}. As previously discussed in Figures \ref{rmse_total} and \ref{rmse_siv_month}, the most substantial RMSE improvement occurs with 20 \% of training samples (Fig. \ref{rmse_siv_map}h). When 20 \% of training samples are used, the SIV RMSE improvement by PINN reaches up to 10 \% near the central Arctic and northern Canadian Archipelago (Fig. \ref{rmse_siv_map}g). When 50 \% of training samples are used, RMSE reductions are observed across the central Arctic, but it is not statistically significant, with less than 5 \% SIV improvement in most regions (Fig. \ref{rmse_siv_map}h). Similarly, when 100 \% samples are used to train the model, $<$5 \% of SIV RMSE reduction is observed across the Arctic Ocean except central Arctic (Fig. \ref{rmse_siv_map}i). 


\begin{figure}
\centerline{\includegraphics[width=1.0\columnwidth]{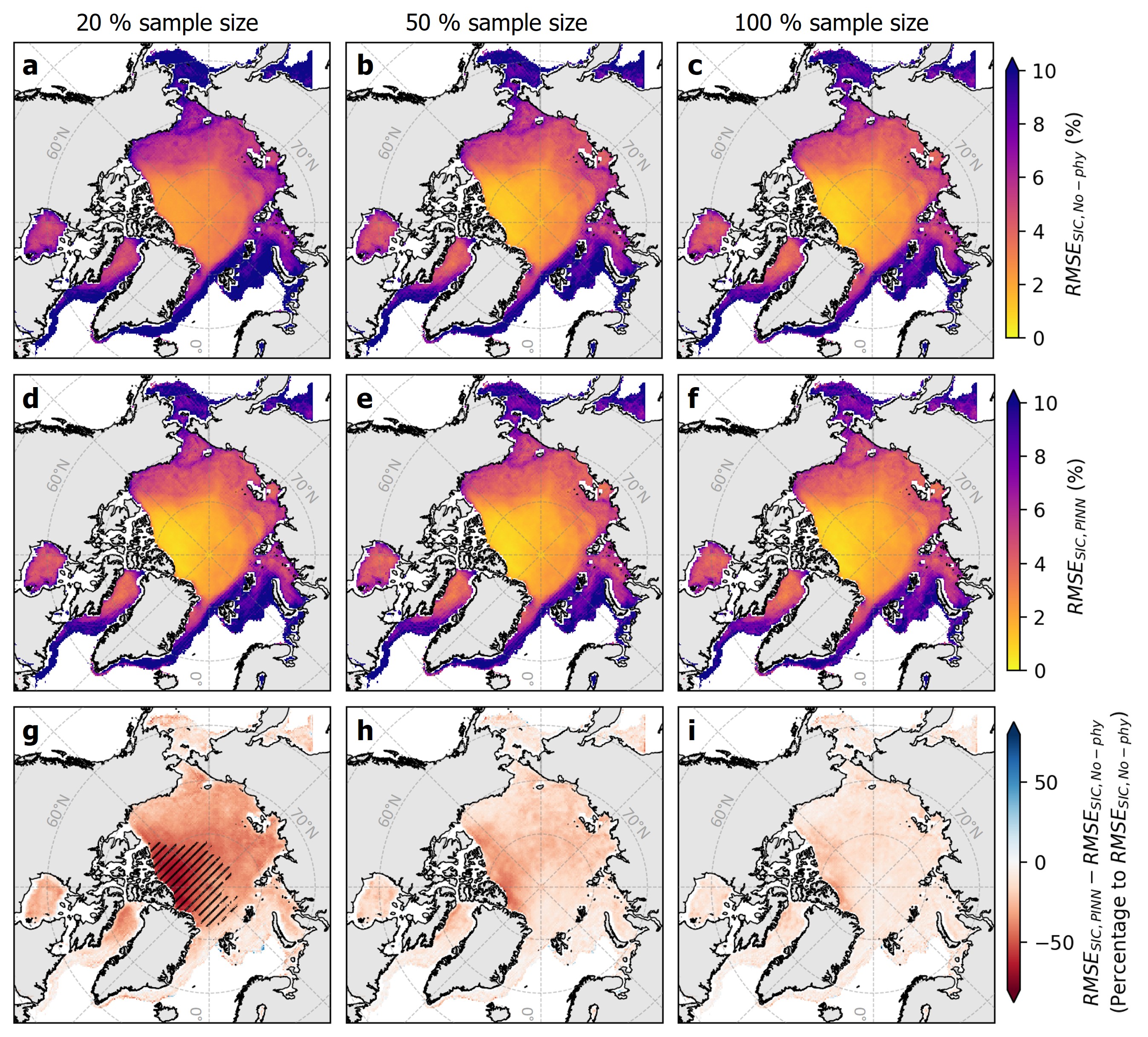}}
\caption{The SIC RMSE map of the No-Phy model for (a) 20 \% training samples, (b) 50 \% training samples, and (c) 100 \% training samples. The SIC RMSE map of the PINN model ($\lambda_{sat}=0.2$ and $\lambda_{therm}=0.2$) for (d) 20 \% training samples, (e) 50 \% training samples, and (f) 100 \% training samples. The relative SIC RMSE difference map between PINN and No-Phy model for (g) 20 \% training samples, (h) 50 \% training samples, and (i) 100 \% training samples. On the bottom panels of the RMSE differences, the reddish color indicates the improvement in SIC RMSE by PINN over the No-Phy model (i.e., reduction in RMSE).}
\label{rmse_sic_map}
\end{figure}

The spatial distribution of SIC RMSE improvement (Fig. \ref{rmse_sic_map}) exhibits a different pattern from SIV. Above all, SIC RMSE improvement is observed across the Arctic Ocean. The PINN approach plays a particularly important role in improving SIC RMSE when only 20 \% of training samples are used. When only 20 \% of training samples are used to train the model, the baseline SIC RMSE of the \emph{No-Phy} model exceeds 4 \% across the Arctic Ocean (Fig. \ref{rmse_sic_map}a). However, by adopting PINN, SIC RMSE decreases significantly by up to 50 \% in the central Arctic (Fig. \ref{rmse_sic_map}g). The improvement of SIC RMSE by PINN is attenuated when more training samples are used. Although SIC RMSE improvement is not significant with 50 \% and 100 \% sample experiments, the effectiveness of PINN still appears widespread over the central Arctic with up to $\sim$20 \% (Fig. \ref{rmse_sic_map}h and i).



\subsection{Implications to future Arctic sea ice prediction}

Given that many studies have shown the Arctic sea ice entered a new state in the 2010s with a dramatic loss of sea ice cover \citep{Stroeve2018, Meier2014, Guemas2016, Doscher2014}, historical satellite-based sea ice observations spanning over more than 40 years since the 1980s may no longer fully represent the recent and future non-stationary sea ice state in the Arctic. Moreover, considering that the Arctic Ocean is projected to be nearly ice-free during summer as early as 2030-2040 \citep{Massonnet2012, Overland2013, Årthun2021}, the future sea ice dynamics in the Arctic Ocean will likely be different from those in historical observations. Hence, this raises concerns about the reliability of purely data-driven deep learning models trained solely on historical records, as such models may struggle to generalize to recent and future non-stationary sea ice conditions under rapid climate change. However, our PINN strategy provides a more robust alternative to fully data-driven learning by integrating physical principles directly into the learning process, thereby reducing reliance on historical data alone and enhancing model generalizability in a non-stationary climate regime.

In this study, we train our PINN models with 2009-2015 data and evaluate them with 2016-2022 data. Compared to the training seven years, the test seven years are characterized as fast-moving sea ice and lower SIC (Fig. \ref{mean_SIC_SIV}). Despite these differences, our PINN indeed shows significant improvement in both SIV and SIC RMSEs compared to the fully data-driven \emph{No-Phy} model. We highlight that our PINN models maintain consistent performance regardless of the number of training samples, whereas the performance of \emph{No-Phy} is dependent on the number of training samples (Figs. \ref{rmse_siv_map} and \ref{rmse_sic_map}). This suggests the potential of the PINN framework to guarantee significant predictive performance and generalizability even when the training data do not fully represent unseen sea ice conditions. Moreover, considering that the significant improvements by PINN are observed near multi-year ice (MYI) regions in the central Arctic and Canadian Archipelago (Figs. \ref{rmse_siv_map} and \ref{rmse_sic_map}), our findings suggest that PINNs can contribute to obtaining consistent model performance despite the ongoing reduction of MYI coverage.

\section{Conclusion}

In this study, we propose a physics-informed learning strategy to improve the fidelity of the existing deep learning model for the prediction of daily sea ice velocity (SIV) and sea ice concentration (SIC) retrieved from spaceborne remote sensing data. Our physics-informed learning strategy is achieved through two pathways: (1) design additional physics-informed loss functions that regularize SIV values based on SIC values and regularize SIC values based on the assumption of daily thermodynamic ice growth and melting; (2) insert sigmoid activation function to restrict the output SIC values into the range of 0 to 1. We implement extensive experiments by employing the Hierarchical information-sharing U-net (HIS-Unet), which predicts SIV and SIC through a series of information sharing between SIV and SIC branches. In order to investigate the impact of training samples on the model robustness, we train the physics-informed neural network (PINN) and normal data-driven neural network (\emph{No-Phy}) separately with three different training sample ratios (100 \%, 50 \%, and 20 \%) and four different weights to the physics-informed loss terms (0.0, 0.2, 1.0, and 0.5).

The results exhibit that the physics-informed learning strategy improves both SIC and SIV predictions. In particular, the improvement of SIC prediction by PINN is more obvious than SIV with a small number of samples. Physics-informed learning strategy helps achieve the benefits of HIS-Unet even in the case of a lack of sufficient training datasets. Therefore, this physics-informed learning strategy can contribute to improving the performance of deep learning models for sea ice prediction, even if the past sea ice data cannot fully represent the future non-stationary sea ice conditions affected by rapid climate change. Additionally, this strategy can also be easily applied for a multi-day forecast of sea ice conditions combined with recurrent network architectures, such as LSTM.

\section*{Software and data availability}
The codes that were used for the prediction of sea ice concentration and velocity using Python language (version 3.11) based on PyTorch library can be found in Github: https://github.com/BinaLab/PINN\_seaice. This repository was created by Younghyun Koo (e-mail: kooala317@gmail.com) in 2023 and contains program codes (111 MB) and training data (3.41 GB). The authors' experimental environment was as follows:
\begin{itemize}
\item OS: Windows 11 Pro
\item CPU: Intel(R) Core(TM) i7-11700F
\item RAM: 16.0 GB
\item GPU: NVIDIA GeForce RTX 3070
\end{itemize}
The sea ice velocity and concentration data can be downloaded free of charge from NSIDC \citep{NSIDC, Meier2021}, and the ERA5 weather data can be downloaded free of charge from the Copernicus Climate Data Store \citep{ERA5}.

\section*{CRediT authorship contribution statement}
\textbf{Younghyun Koo}: Software, Validation, Formal analysis, Data curation, Visualization, Writing - Original Draft. \textbf{Maryam Rahnemoonfar}: Conceptualization, Methodology, Writing - Review \& Editing, Supervision, Funding acquisition.

\section*{Acknowledgments}
This work is supported by the U.S. National Science Foundation (NSF) BIGDATA (IIS-1838230, IIS-2308649) and NSF Leadership Class Computing (OAC-2139536) awards.

\appendix

\section{Arctic SIC and SIV conditions from 2016 to 2022}\label{appendix}

\begin{figure}
\centerline{\includegraphics[width=1.0\columnwidth]{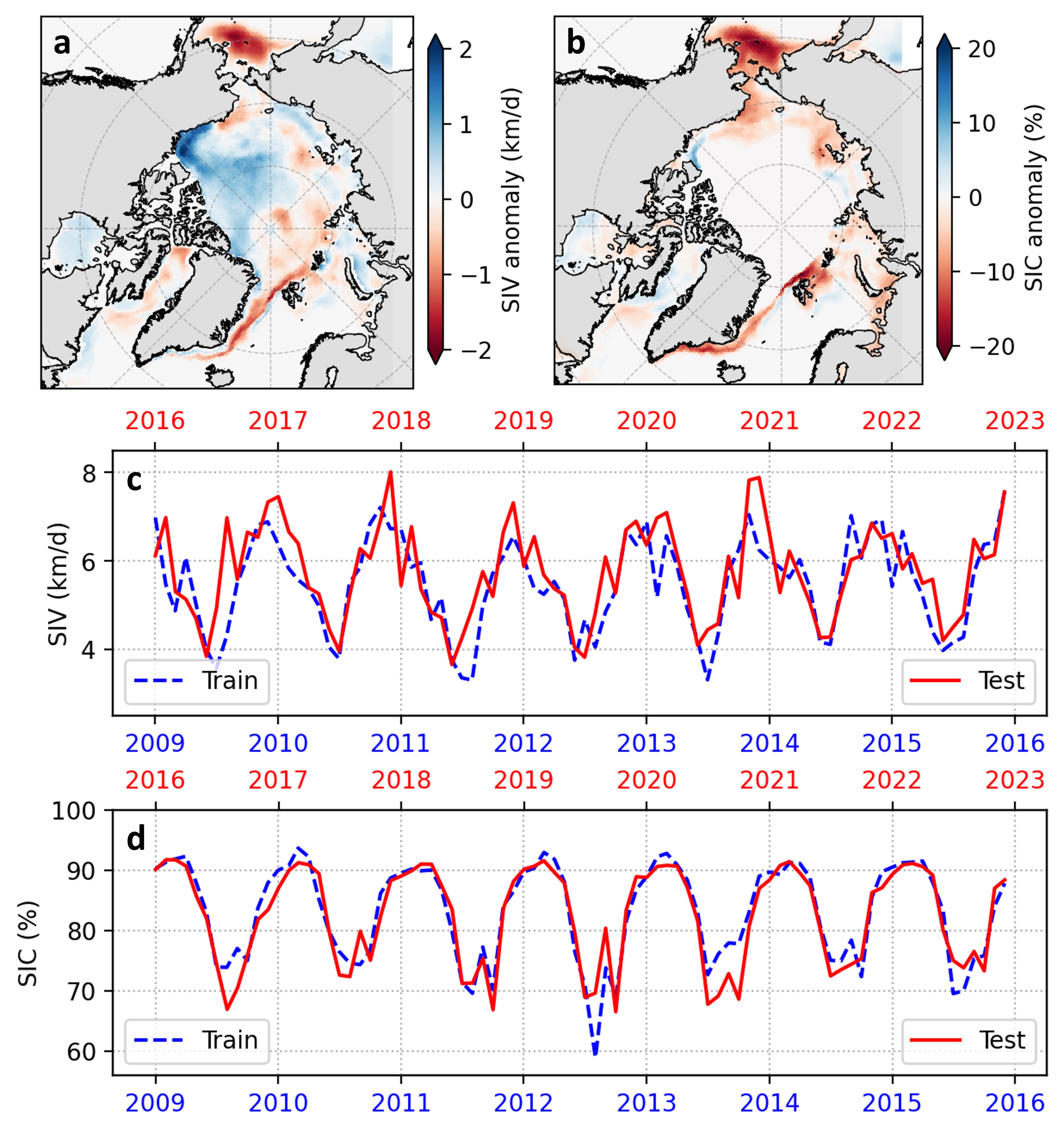}}
\caption{Maps of (a) SIV anomaly and (b) SIC anomaly in the test years (2016-2022) compared to the training years (2009-2015). Monthly mean (c) SIV and (d) SIC in training years and test years.}
\label{mean_SIC_SIV}
\end{figure}

Figures \ref{mean_SIC_SIV}a and b show the average SIV and SIC, respectively, over the Arctic Ocean in 2022. The Fram Strait in eastern Greenland shows the fastest ice movement. Figures \ref{mean_SIC_SIV}c and d show the monthly mean SIV and SIC, respectively, in 2022. SIV decreased from January to June and increased again thereafter. SIC was consistent to 90 \% from January to May, decreased to $\sim$70 \% until September, and increased again from October.

\bibliographystyle{elsarticle-harv} 
\bibliography{reference}

\begin{thebibliography}{75}
\expandafter\ifx\csname natexlab\endcsname\relax\def\natexlab#1{#1}\fi
\providecommand{\url}[1]{\texttt{#1}}
\providecommand{\href}[2]{#2}
\providecommand{\path}[1]{#1}
\providecommand{\DOIprefix}{doi:}
\providecommand{\ArXivprefix}{arXiv:}
\providecommand{\URLprefix}{URL: }
\providecommand{\Pubmedprefix}{pmid:}
\providecommand{\doi}[1]{\href{http://dx.doi.org/#1}{\path{#1}}}
\providecommand{\Pubmed}[1]{\href{pmid:#1}{\path{#1}}}
\providecommand{\bibinfo}[2]{#2}
\ifx\xfnm\relax \def\xfnm[#1]{\unskip,\space#1}\fi
\bibitem[{Andersson et~al.(2021)Andersson, Hosking, Pérez-Ortiz, Paige, Elliott, Russell, Law, Jones, Wilkinson, Phillips, Byrne, Tietsche, Sarojini, Blanchard-Wrigglesworth, Aksenov, Downie and Shuckburgh}]{Andersson2021}
\bibinfo{author}{Andersson, T.R.}, \bibinfo{author}{Hosking, J.S.}, \bibinfo{author}{Pérez-Ortiz, M.}, \bibinfo{author}{Paige, B.}, \bibinfo{author}{Elliott, A.}, \bibinfo{author}{Russell, C.}, \bibinfo{author}{Law, S.}, \bibinfo{author}{Jones, D.C.}, \bibinfo{author}{Wilkinson, J.}, \bibinfo{author}{Phillips, T.}, \bibinfo{author}{Byrne, J.}, \bibinfo{author}{Tietsche, S.}, \bibinfo{author}{Sarojini, B.B.}, \bibinfo{author}{Blanchard-Wrigglesworth, E.}, \bibinfo{author}{Aksenov, Y.}, \bibinfo{author}{Downie, R.}, \bibinfo{author}{Shuckburgh, E.}, \bibinfo{year}{2021}.
\newblock \bibinfo{title}{Seasonal arctic sea ice forecasting with probabilistic deep learning}.
\newblock \bibinfo{journal}{Nature Communications} \bibinfo{volume}{12}.
\bibitem[{Blanchard-Wrigglesworth et~al.(2015)Blanchard-Wrigglesworth, Cullather, Wang, Zhang and Bitz}]{Wrigglesworth2015}
\bibinfo{author}{Blanchard-Wrigglesworth, E.}, \bibinfo{author}{Cullather, R.I.}, \bibinfo{author}{Wang, W.}, \bibinfo{author}{Zhang, J.}, \bibinfo{author}{Bitz, C.M.}, \bibinfo{year}{2015}.
\newblock \bibinfo{title}{{Model forecast skill and sensitivity to initial conditions in the seasonal Sea Ice Outlook}}.
\newblock \bibinfo{journal}{Geophysical Research Letters} \bibinfo{volume}{42}, \bibinfo{pages}{8042--8048}.
\newblock \DOIprefix\doi{10.1002/2015GL065860}.
\bibitem[{Blockley et~al.(2020)Blockley, Vancoppenolle, Hunke, Bitz, Feltham, Lemieux, Losch, Maisonnave, Notz, Rampal, Tietsche, Tremblay, Turner, Massonnet, Ólason, Roberts, Aksenov, Fichefet, Garric, Iovino, Madec, Rousset, y~Melia and Schroeder}]{Blockley2020}
\bibinfo{author}{Blockley, E.}, \bibinfo{author}{Vancoppenolle, M.}, \bibinfo{author}{Hunke, E.}, \bibinfo{author}{Bitz, C.}, \bibinfo{author}{Feltham, D.}, \bibinfo{author}{Lemieux, J.F.}, \bibinfo{author}{Losch, M.}, \bibinfo{author}{Maisonnave, E.}, \bibinfo{author}{Notz, D.}, \bibinfo{author}{Rampal, P.}, \bibinfo{author}{Tietsche, S.}, \bibinfo{author}{Tremblay, B.}, \bibinfo{author}{Turner, A.}, \bibinfo{author}{Massonnet, F.}, \bibinfo{author}{Ólason, E.}, \bibinfo{author}{Roberts, A.}, \bibinfo{author}{Aksenov, Y.}, \bibinfo{author}{Fichefet, T.}, \bibinfo{author}{Garric, G.}, \bibinfo{author}{Iovino, D.}, \bibinfo{author}{Madec, G.}, \bibinfo{author}{Rousset, C.}, \bibinfo{author}{y~Melia, D.S.}, \bibinfo{author}{Schroeder, D.}, \bibinfo{year}{2020}.
\newblock \bibinfo{title}{The future of sea ice modeling: Where do we go from here?}
\newblock \bibinfo{journal}{Bulletin of the American Meteorological Society} \bibinfo{volume}{101}, \bibinfo{pages}{E1304 -- E1311}.
\newblock \DOIprefix\doi{10.1175/BAMS-D-20-0073.1}.
\bibitem[{Budikova(2009)}]{BUDIKOVA2009}
\bibinfo{author}{Budikova, D.}, \bibinfo{year}{2009}.
\newblock \bibinfo{title}{{Role of Arctic sea ice in global atmospheric circulation: A review}}.
\newblock \bibinfo{journal}{Global and Planetary Change} \bibinfo{volume}{68}, \bibinfo{pages}{149--163}.
\newblock \DOIprefix\doi{10.1016/j.gloplacha.2009.04.001}.
\bibitem[{Cavalieri et~al.(1984)Cavalieri, Gloersen and Campbell}]{NASA_team}
\bibinfo{author}{Cavalieri, D.J.}, \bibinfo{author}{Gloersen, P.}, \bibinfo{author}{Campbell, W.J.}, \bibinfo{year}{1984}.
\newblock \bibinfo{title}{{Determination of sea ice parameters with the NIMBUS 7 SMMR}}.
\newblock \bibinfo{journal}{Journal of Geophysical Research: Atmospheres} \bibinfo{volume}{89}, \bibinfo{pages}{5355--5369}.
\newblock \DOIprefix\doi{10.1029/JD089iD04p05355}.
\bibitem[{Cheng et~al.(2024)Cheng, Morlighem and Francis}]{Cheng2024}
\bibinfo{author}{Cheng, G.}, \bibinfo{author}{Morlighem, M.}, \bibinfo{author}{Francis, S.}, \bibinfo{year}{2024}.
\newblock \bibinfo{title}{{Forward and Inverse Modeling of Ice Sheet Flow Using Physics-Informed Neural Networks: Application to Helheim Glacier, Greenland}}.
\newblock \bibinfo{journal}{Journal of Geophysical Research: Machine Learning and Computation} \bibinfo{volume}{1}, \bibinfo{pages}{e2024JH000169}.
\newblock \DOIprefix\doi{10.1029/2024JH000169}.
\bibitem[{Comiso(1986)}]{NASA_BT}
\bibinfo{author}{Comiso, J.C.}, \bibinfo{year}{1986}.
\newblock \bibinfo{title}{{Characteristics of Arctic winter sea ice from satellite multispectral microwave observations}}.
\newblock \bibinfo{journal}{Journal of Geophysical Research: Oceans} \bibinfo{volume}{91}, \bibinfo{pages}{975--994}.
\newblock \DOIprefix\doi{10.1029/JC091iC01p00975}.
\bibitem[{Comiso et~al.(1997)Comiso, Cavalieri, Parkinson and Gloersen}]{Comiso1997}
\bibinfo{author}{Comiso, J.C.}, \bibinfo{author}{Cavalieri, D.J.}, \bibinfo{author}{Parkinson, C.L.}, \bibinfo{author}{Gloersen, P.}, \bibinfo{year}{1997}.
\newblock \bibinfo{title}{Passive microwave algorithms for sea ice concentration: A comparison of two techniques}.
\newblock \bibinfo{journal}{Remote Sensing of Environment} \bibinfo{volume}{60}, \bibinfo{pages}{357--384}.
\newblock \DOIprefix\doi{10.1016/S0034-4257(96)00220-9}.
\bibitem[{D\"oscher et~al.(2014)D\"oscher, Vihma and Maksimovich}]{Doscher2014}
\bibinfo{author}{D\"oscher, R.}, \bibinfo{author}{Vihma, T.}, \bibinfo{author}{Maksimovich, E.}, \bibinfo{year}{2014}.
\newblock \bibinfo{title}{Recent advances in understanding the {Arctic} climate system state and change from a sea ice perspective: a review}.
\newblock \bibinfo{journal}{Atmospheric Chemistry and Physics} \bibinfo{volume}{14}, \bibinfo{pages}{13571--13600}.
\newblock \DOIprefix\doi{10.5194/acp-14-13571-2014}.
\bibitem[{Flato(2004)}]{flato2004}
\bibinfo{author}{Flato, G.M.}, \bibinfo{year}{2004}.
\newblock \bibinfo{title}{Sea-ice modelling}, in: \bibinfo{editor}{Bamber, J.L.}, \bibinfo{editor}{Payne, A.J.} (Eds.), \bibinfo{booktitle}{Mass Balance of the Cryosphere: Observations and Modelling of Contemporary and Future Changes}. \bibinfo{publisher}{Cambridge University Press}, p. \bibinfo{pages}{367–390}.
\newblock \DOIprefix\doi{10.1017/CBO9780511535659.011}.
\bibitem[{Fritzner et~al.(2020)Fritzner, Graversen and Christensen}]{Fritzner2020}
\bibinfo{author}{Fritzner, S.}, \bibinfo{author}{Graversen, R.}, \bibinfo{author}{Christensen, K.H.}, \bibinfo{year}{2020}.
\newblock \bibinfo{title}{Assessment of high-resolution dynamical and machine learning models for prediction of sea ice concentration in a regional application}.
\newblock \bibinfo{journal}{Journal of Geophysical Research: Oceans} \bibinfo{volume}{125}, \bibinfo{pages}{e2020JC016277}.
\newblock \DOIprefix\doi{10.1029/2020JC016277}.
\bibitem[{Fritzner et~al.(2019)Fritzner, Graversen, Christensen, Rostosky and Wang}]{Fritzner2019}
\bibinfo{author}{Fritzner, S.}, \bibinfo{author}{Graversen, R.}, \bibinfo{author}{Christensen, K.H.}, \bibinfo{author}{Rostosky, P.}, \bibinfo{author}{Wang, K.}, \bibinfo{year}{2019}.
\newblock \bibinfo{title}{Impact of assimilating sea ice concentration, sea ice thickness and snow depth in a coupled ocean--sea ice modelling system}.
\newblock \bibinfo{journal}{The Cryosphere} \bibinfo{volume}{13}, \bibinfo{pages}{491--509}.
\newblock \DOIprefix\doi{10.5194/tc-13-491-2019}.
\bibitem[{Grigoryev et~al.(2022)Grigoryev, Verezemskaya, Krinitskiy, Anikin, Gavrikov, Trofimov, Balabin, Shpilman, Eremchenko, Gulev, Burnaev and Vanovskiy}]{Grigoryev2022}
\bibinfo{author}{Grigoryev, T.}, \bibinfo{author}{Verezemskaya, P.}, \bibinfo{author}{Krinitskiy, M.}, \bibinfo{author}{Anikin, N.}, \bibinfo{author}{Gavrikov, A.}, \bibinfo{author}{Trofimov, I.}, \bibinfo{author}{Balabin, N.}, \bibinfo{author}{Shpilman, A.}, \bibinfo{author}{Eremchenko, A.}, \bibinfo{author}{Gulev, S.}, \bibinfo{author}{Burnaev, E.}, \bibinfo{author}{Vanovskiy, V.}, \bibinfo{year}{2022}.
\newblock \bibinfo{title}{Data-driven short-term daily operational sea ice regional forecasting}.
\newblock \bibinfo{journal}{Remote Sensing} \bibinfo{volume}{14}.
\newblock \DOIprefix\doi{10.3390/rs14225837}.
\bibitem[{Guemas et~al.(2016)Guemas, Blanchard-Wrigglesworth, Chevallier, Day, Déqué, Doblas-Reyes, Fučkar, Germe, Hawkins, Keeley, Koenigk, Salas~y Mélia and Tietsche}]{Guemas2016}
\bibinfo{author}{Guemas, V.}, \bibinfo{author}{Blanchard-Wrigglesworth, E.}, \bibinfo{author}{Chevallier, M.}, \bibinfo{author}{Day, J.J.}, \bibinfo{author}{Déqué, M.}, \bibinfo{author}{Doblas-Reyes, F.J.}, \bibinfo{author}{Fučkar, N.S.}, \bibinfo{author}{Germe, A.}, \bibinfo{author}{Hawkins, E.}, \bibinfo{author}{Keeley, S.}, \bibinfo{author}{Koenigk, T.}, \bibinfo{author}{Salas~y Mélia, D.}, \bibinfo{author}{Tietsche, S.}, \bibinfo{year}{2016}.
\newblock \bibinfo{title}{A review on {Arctic} sea-ice predictability and prediction on seasonal to decadal time-scales}.
\newblock \bibinfo{journal}{Quarterly Journal of the Royal Meteorological Society} \bibinfo{volume}{142}, \bibinfo{pages}{546--561}.
\newblock \DOIprefix\doi{10.1002/qj.2401}.
\bibitem[{Hao et~al.(2023)Hao, Liu, Zhang, Ying, Feng, Su and Zhu}]{Hao2023_PIML}
\bibinfo{author}{Hao, Z.}, \bibinfo{author}{Liu, S.}, \bibinfo{author}{Zhang, Y.}, \bibinfo{author}{Ying, C.}, \bibinfo{author}{Feng, Y.}, \bibinfo{author}{Su, H.}, \bibinfo{author}{Zhu, J.}, \bibinfo{year}{2023}.
\newblock \bibinfo{title}{Physics-informed machine learning: A survey on problems, methods and applications}.
\newblock \DOIprefix\doi{10.48550/arXiv.2211.08064}.
\bibitem[{Haumann et~al.(2016)Haumann, Gruber, M{\"u}nnich, Frenger and Kern}]{haumann2016sea}
\bibinfo{author}{Haumann, F.A.}, \bibinfo{author}{Gruber, N.}, \bibinfo{author}{M{\"u}nnich, M.}, \bibinfo{author}{Frenger, I.}, \bibinfo{author}{Kern, S.}, \bibinfo{year}{2016}.
\newblock \bibinfo{title}{{Sea-ice transport driving Southern Ocean salinity and its recent trends}}.
\newblock \bibinfo{journal}{Nature} \bibinfo{volume}{537}, \bibinfo{pages}{89--92}.
\newblock \DOIprefix\doi{10.1038/nature19101}.
\bibitem[{He et~al.(2023)He, Perego, Howard, Karniadakis and Stinis}]{HE2023}
\bibinfo{author}{He, Q.}, \bibinfo{author}{Perego, M.}, \bibinfo{author}{Howard, A.A.}, \bibinfo{author}{Karniadakis, G.E.}, \bibinfo{author}{Stinis, P.}, \bibinfo{year}{2023}.
\newblock \bibinfo{title}{A hybrid deep neural operator/finite element method for ice-sheet modeling}.
\newblock \bibinfo{journal}{Journal of Computational Physics} \bibinfo{volume}{492}, \bibinfo{pages}{112428}.
\newblock \DOIprefix\doi{10.1016/j.jcp.2023.112428}.
\bibitem[{Hersbach et~al.(2020)Hersbach, Bell, Berrisford, Hirahara, Horányi, Muñoz-Sabater, Nicolas, Peubey, Radu, Schepers, Simmons, Soci, Abdalla, Abellan, Balsamo, Bechtold, Biavati, Bidlot, Bonavita, De~Chiara, Dahlgren, Dee, Diamantakis, Dragani, Flemming, Forbes, Fuentes, Geer, Haimberger, Healy, Hogan, Hólm, Janisková, Keeley, Laloyaux, Lopez, Lupu, Radnoti, de~Rosnay, Rozum, Vamborg, Villaume and Thépaut}]{ERA5}
\bibinfo{author}{Hersbach, H.}, \bibinfo{author}{Bell, B.}, \bibinfo{author}{Berrisford, P.}, \bibinfo{author}{Hirahara, S.}, \bibinfo{author}{Horányi, A.}, \bibinfo{author}{Muñoz-Sabater, J.}, \bibinfo{author}{Nicolas, J.}, \bibinfo{author}{Peubey, C.}, \bibinfo{author}{Radu, R.}, \bibinfo{author}{Schepers, D.}, \bibinfo{author}{Simmons, A.}, \bibinfo{author}{Soci, C.}, \bibinfo{author}{Abdalla, S.}, \bibinfo{author}{Abellan, X.}, \bibinfo{author}{Balsamo, G.}, \bibinfo{author}{Bechtold, P.}, \bibinfo{author}{Biavati, G.}, \bibinfo{author}{Bidlot, J.}, \bibinfo{author}{Bonavita, M.}, \bibinfo{author}{De~Chiara, G.}, \bibinfo{author}{Dahlgren, P.}, \bibinfo{author}{Dee, D.}, \bibinfo{author}{Diamantakis, M.}, \bibinfo{author}{Dragani, R.}, \bibinfo{author}{Flemming, J.}, \bibinfo{author}{Forbes, R.}, \bibinfo{author}{Fuentes, M.}, \bibinfo{author}{Geer, A.}, \bibinfo{author}{Haimberger, L.}, \bibinfo{author}{Healy, S.}, \bibinfo{author}{Hogan, R.J.}, \bibinfo{author}{Hólm, E.}, \bibinfo{author}{Janisková,
  M.}, \bibinfo{author}{Keeley, S.}, \bibinfo{author}{Laloyaux, P.}, \bibinfo{author}{Lopez, P.}, \bibinfo{author}{Lupu, C.}, \bibinfo{author}{Radnoti, G.}, \bibinfo{author}{de~Rosnay, P.}, \bibinfo{author}{Rozum, I.}, \bibinfo{author}{Vamborg, F.}, \bibinfo{author}{Villaume, S.}, \bibinfo{author}{Thépaut, J.N.}, \bibinfo{year}{2020}.
\newblock \bibinfo{title}{{The ERA5 global reanalysis}}.
\newblock \bibinfo{journal}{Quarterly Journal of the Royal Meteorological Society} \bibinfo{volume}{146}, \bibinfo{pages}{1999--2049}.
\newblock \DOIprefix\doi{10.1002/qj.3803}.
\bibitem[{Hibler(1979)}]{Hibler1979}
\bibinfo{author}{Hibler, W.D.}, \bibinfo{year}{1979}.
\newblock \bibinfo{title}{A dynamic thermodynamic sea ice model}.
\newblock \bibinfo{journal}{Journal of Physical Oceanography} \bibinfo{volume}{9}, \bibinfo{pages}{815 -- 846}.
\newblock \DOIprefix\doi{10.1175/1520-0485(1979)009$<$0815:ADTSIM$>$2.0.CO;2}.
\bibitem[{Hirst(1999)}]{Hirst1999}
\bibinfo{author}{Hirst, A.C.}, \bibinfo{year}{1999}.
\newblock \bibinfo{title}{{The Southern Ocean response to global warming in the CSIRO coupled ocean-atmosphere model}}.
\newblock \bibinfo{journal}{Environmental Modelling \& Software} \bibinfo{volume}{14}, \bibinfo{pages}{227--241}.
\newblock \DOIprefix\doi{10.1016/S1364-8152(98)00075-9}.
\bibitem[{Hoffman et~al.(2023)Hoffman, Mazloff, Gille, Giglio, Bitz, Heimbach and Matsuyoshi}]{Hoffman2023}
\bibinfo{author}{Hoffman, L.}, \bibinfo{author}{Mazloff, M.R.}, \bibinfo{author}{Gille, S.T.}, \bibinfo{author}{Giglio, D.}, \bibinfo{author}{Bitz, C.M.}, \bibinfo{author}{Heimbach, P.}, \bibinfo{author}{Matsuyoshi, K.}, \bibinfo{year}{2023}.
\newblock \bibinfo{title}{{Machine learning for daily forecasts of Arctic sea-ice motion: an attribution assessment of model predictive skill}}.
\newblock \bibinfo{journal}{Artificial Intelligence for the Earth Systems} , \bibinfo{pages}{1 -- 45}\DOIprefix\doi{10.1175/AIES-D-23-0004.1}.
\bibitem[{Holland and Kwok(2012)}]{Holland2012}
\bibinfo{author}{Holland, P.R.}, \bibinfo{author}{Kwok, R.}, \bibinfo{year}{2012}.
\newblock \bibinfo{title}{Wind-driven trends in {Antarctic} sea-ice drift}.
\newblock \bibinfo{journal}{Nature Geoscience} \bibinfo{volume}{5}, \bibinfo{pages}{872--875}.
\newblock \DOIprefix\doi{10.1038/ngeo1627}.
\bibitem[{Hunke and Dukowicz(1997)}]{Hunke1997}
\bibinfo{author}{Hunke, E.C.}, \bibinfo{author}{Dukowicz, J.K.}, \bibinfo{year}{1997}.
\newblock \bibinfo{title}{An elastic–viscous–plastic model for sea ice dynamics}.
\newblock \bibinfo{journal}{Journal of Physical Oceanography} \bibinfo{volume}{27}, \bibinfo{pages}{1849 -- 1867}.
\newblock \DOIprefix\doi{10.1175/1520-0485(1997)027$<$1849:AEVPMF$>$2.0.CO;2}.
\bibitem[{Hunke et~al.(2010)Hunke, Lipscomb and Turner}]{Hunke2010}
\bibinfo{author}{Hunke, E.C.}, \bibinfo{author}{Lipscomb, W.H.}, \bibinfo{author}{Turner, A.K.}, \bibinfo{year}{2010}.
\newblock \bibinfo{title}{Sea-ice models for climate study: retrospective and new directions}.
\newblock \bibinfo{journal}{Journal of Glaciology} \bibinfo{volume}{56}, \bibinfo{pages}{1162–1172}.
\newblock \DOIprefix\doi{10.3189/002214311796406095}.
\bibitem[{Ivanova et~al.(2015)Ivanova, Pedersen, Tonboe, Kern, Heygster, Lavergne, S{\o}rensen, Saldo, Dybkj{\ae}r, Brucker and Shokr}]{Ivanova2015}
\bibinfo{author}{Ivanova, N.}, \bibinfo{author}{Pedersen, L.T.}, \bibinfo{author}{Tonboe, R.T.}, \bibinfo{author}{Kern, S.}, \bibinfo{author}{Heygster, G.}, \bibinfo{author}{Lavergne, T.}, \bibinfo{author}{S{\o}rensen, A.}, \bibinfo{author}{Saldo, R.}, \bibinfo{author}{Dybkj{\ae}r, G.}, \bibinfo{author}{Brucker, L.}, \bibinfo{author}{Shokr, M.}, \bibinfo{year}{2015}.
\newblock \bibinfo{title}{Inter-comparison and evaluation of sea ice algorithms: towards further identification of challenges and optimal approach using passive microwave observations}.
\newblock \bibinfo{journal}{The Cryosphere} \bibinfo{volume}{9}, \bibinfo{pages}{1797--1817}.
\newblock \DOIprefix\doi{10.5194/tc-9-1797-2015}.
\bibitem[{Iwasaki and Lai(2023)}]{Iwasaki2023}
\bibinfo{author}{Iwasaki, Y.}, \bibinfo{author}{Lai, C.Y.}, \bibinfo{year}{2023}.
\newblock \bibinfo{title}{One-dimensional ice shelf hardness inversion: Clustering behavior and collocation resampling in physics-informed neural networks}.
\newblock \bibinfo{journal}{Journal of Computational Physics} \bibinfo{volume}{492}, \bibinfo{pages}{112435}.
\newblock \DOIprefix\doi{10.1016/j.jcp.2023.112435}.
\bibitem[{Jiang et~al.(2023)Jiang, Qiao, Su, Li, Meng, Wu, Quan, Wang and Wang}]{JIANG2023_EMS_CNN}
\bibinfo{author}{Jiang, W.}, \bibinfo{author}{Qiao, Y.}, \bibinfo{author}{Su, G.}, \bibinfo{author}{Li, X.}, \bibinfo{author}{Meng, Q.}, \bibinfo{author}{Wu, H.}, \bibinfo{author}{Quan, W.}, \bibinfo{author}{Wang, J.}, \bibinfo{author}{Wang, F.}, \bibinfo{year}{2023}.
\newblock \bibinfo{title}{{WFNet: A hierarchical convolutional neural network for wildfire spread prediction}}.
\newblock \bibinfo{journal}{Environmental Modelling \& Software} \bibinfo{volume}{170}, \bibinfo{pages}{105841}.
\newblock \DOIprefix\doi{10.1016/j.envsoft.2023.105841}.
\bibitem[{Jouvet and Cordonnier(2023)}]{Jouvet2023}
\bibinfo{author}{Jouvet, G.}, \bibinfo{author}{Cordonnier, G.}, \bibinfo{year}{2023}.
\newblock \bibinfo{title}{Ice-flow model emulator based on physics-informed deep learning}.
\newblock \bibinfo{journal}{Journal of Glaciology} , \bibinfo{pages}{1–15}\DOIprefix\doi{10.1017/jog.2023.73}.
\bibitem[{Karniadakis et~al.(2021)Karniadakis, Kevrekidis, Lu, Perdikaris, Wang and Yang}]{Karniadakis2021}
\bibinfo{author}{Karniadakis, G.E.}, \bibinfo{author}{Kevrekidis, I.G.}, \bibinfo{author}{Lu, L.}, \bibinfo{author}{Perdikaris, P.}, \bibinfo{author}{Wang, S.}, \bibinfo{author}{Yang, L.}, \bibinfo{year}{2021}.
\newblock \bibinfo{title}{Physics-informed machine learning}.
\newblock \bibinfo{journal}{Nature Reviews Physics} \bibinfo{volume}{3}.
\newblock \DOIprefix\doi{10.1038/s42254-021-00314-5}.
\bibitem[{Kern et~al.(2020)Kern, Lavergne, Notz, Pedersen and Tonboe}]{Kern2020}
\bibinfo{author}{Kern, S.}, \bibinfo{author}{Lavergne, T.}, \bibinfo{author}{Notz, D.}, \bibinfo{author}{Pedersen, L.T.}, \bibinfo{author}{Tonboe, R.}, \bibinfo{year}{2020}.
\newblock \bibinfo{title}{Satellite passive microwave sea-ice concentration data set inter-comparison for arctic summer conditions}.
\newblock \bibinfo{journal}{The Cryosphere} \bibinfo{volume}{14}, \bibinfo{pages}{2469--2493}.
\newblock \DOIprefix\doi{10.5194/tc-14-2469-2020}.
\bibitem[{Kim et~al.(2020)Kim, Kim, Han, Lee and Im}]{Kim2020}
\bibinfo{author}{Kim, Y.J.}, \bibinfo{author}{Kim, H.C.}, \bibinfo{author}{Han, D.}, \bibinfo{author}{Lee, S.}, \bibinfo{author}{Im, J.}, \bibinfo{year}{2020}.
\newblock \bibinfo{title}{{Prediction of monthly Arctic sea ice concentrations using satellite and reanalysis data based on convolutional neural networks}}.
\newblock \bibinfo{journal}{The Cryosphere} \bibinfo{volume}{14}, \bibinfo{pages}{1083--1104}.
\newblock \DOIprefix\doi{10.5194/tc-14-1083-2020}.
\bibitem[{Kingma and Ba(2017)}]{Kingma2015}
\bibinfo{author}{Kingma, D.P.}, \bibinfo{author}{Ba, J.}, \bibinfo{year}{2017}.
\newblock \bibinfo{title}{Adam: A method for stochastic optimization}.
\newblock \DOIprefix\doi{10.48550/arXiv.1412.6980}, \href{http://arxiv.org/abs/1412.6980}{{\tt arXiv:1412.6980}}.
\bibitem[{Koo and Rahnemoonfar(2024)}]{koo2023_hisunet}
\bibinfo{author}{Koo, Y.}, \bibinfo{author}{Rahnemoonfar, M.}, \bibinfo{year}{2024}.
\newblock \bibinfo{title}{Hierarchical information-sharing convolutional neural network for the prediction of {Arctic} sea ice concentration and velocity}.
\newblock \bibinfo{journal}{IEEE Transactions on Geoscience and Remote Sensing} \bibinfo{volume}{62}, \bibinfo{pages}{1--13}.
\newblock \DOIprefix\doi{10.1109/TGRS.2024.3501094}.
\bibitem[{Kwok(2018)}]{Kwok2018}
\bibinfo{author}{Kwok, R.}, \bibinfo{year}{2018}.
\newblock \bibinfo{title}{Arctic sea ice thickness, volume, and multiyear ice coverage: losses and coupled variability (1958-2018)}.
\newblock \bibinfo{journal}{Environmental Research Letters} \bibinfo{volume}{13}, \bibinfo{pages}{105005}.
\newblock \DOIprefix\doi{10.1088/1748-9326/aae3ec}.
\bibitem[{Li et~al.(2023)Li, Gelb and Lee}]{Li2023}
\bibinfo{author}{Li, T.}, \bibinfo{author}{Gelb, A.}, \bibinfo{author}{Lee, Y.}, \bibinfo{year}{2023}.
\newblock \bibinfo{title}{Improving numerical accuracy for the viscous-plastic formulation of sea ice}.
\newblock \bibinfo{journal}{Journal of Computational Physics} \bibinfo{volume}{487}, \bibinfo{pages}{112184}.
\newblock \DOIprefix\doi{10.1016/j.jcp.2023.112184}.
\bibitem[{Liu et~al.(2024)Liu, Wang, Zhang, Zhang, Yan and Liu}]{LIU2024_PINN}
\bibinfo{author}{Liu, Q.}, \bibinfo{author}{Wang, Y.}, \bibinfo{author}{Zhang, R.}, \bibinfo{author}{Zhang, L.}, \bibinfo{author}{Yan, H.}, \bibinfo{author}{Liu, K.}, \bibinfo{year}{2024}.
\newblock \bibinfo{title}{Physics-informed deep convolutional network for combined sea ice concentration and velocity prediction}.
\newblock \bibinfo{journal}{Ocean Engineering} \bibinfo{volume}{313}, \bibinfo{pages}{119440}.
\newblock \DOIprefix\doi{10.1016/j.oceaneng.2024.119440}.
\bibitem[{Liu et~al.(2021)Liu, Zhang, Wang, Yan and Hong}]{Liu2021_daily}
\bibinfo{author}{Liu, Q.}, \bibinfo{author}{Zhang, R.}, \bibinfo{author}{Wang, Y.}, \bibinfo{author}{Yan, H.}, \bibinfo{author}{Hong, M.}, \bibinfo{year}{2021}.
\newblock \bibinfo{title}{Daily prediction of the {Arctic} sea ice concentration using reanalysis data based on a convolutional {LSTM} network}.
\newblock \bibinfo{journal}{Journal of Marine Science and Engineering} \bibinfo{volume}{9}.
\newblock \DOIprefix\doi{10.3390/jmse9030330}.
\bibitem[{Maier et~al.(2023)Maier, Galelli, Razavi, Castelletti, Rizzoli, Athanasiadis, Sànchez-Marrè, Acutis, Wu and Humphrey}]{Maier2023_mythANN}
\bibinfo{author}{Maier, H.R.}, \bibinfo{author}{Galelli, S.}, \bibinfo{author}{Razavi, S.}, \bibinfo{author}{Castelletti, A.}, \bibinfo{author}{Rizzoli, A.}, \bibinfo{author}{Athanasiadis, I.N.}, \bibinfo{author}{Sànchez-Marrè, M.}, \bibinfo{author}{Acutis, M.}, \bibinfo{author}{Wu, W.}, \bibinfo{author}{Humphrey, G.B.}, \bibinfo{year}{2023}.
\newblock \bibinfo{title}{Exploding the myths: An introduction to artificial neural networks for prediction and forecasting}.
\newblock \bibinfo{journal}{Environmental Modelling \& Software} \bibinfo{volume}{167}, \bibinfo{pages}{105776}.
\newblock \DOIprefix\doi{10.1016/j.envsoft.2023.105776}.
\bibitem[{Mallat(2016)}]{Mallat2015}
\bibinfo{author}{Mallat, S.}, \bibinfo{year}{2016}.
\newblock \bibinfo{title}{Understanding deep convolutional networks}.
\newblock \bibinfo{journal}{Philosophical Transactions of the Royal Society A: Mathematical, Physical and Engineering Sciences} \bibinfo{volume}{374}, \bibinfo{pages}{20150203}.
\newblock \DOIprefix\doi{10.1098/rsta.2015.0203}.
\bibitem[{Massonnet et~al.(2012)Massonnet, Fichefet, Goosse, Bitz, Philippon-Berthier, Holland and Barriat}]{Massonnet2012}
\bibinfo{author}{Massonnet, F.}, \bibinfo{author}{Fichefet, T.}, \bibinfo{author}{Goosse, H.}, \bibinfo{author}{Bitz, C.M.}, \bibinfo{author}{Philippon-Berthier, G.}, \bibinfo{author}{Holland, M.M.}, \bibinfo{author}{Barriat, P.Y.}, \bibinfo{year}{2012}.
\newblock \bibinfo{title}{Constraining projections of summer {Arctic} sea ice}.
\newblock \bibinfo{journal}{The Cryosphere} \bibinfo{volume}{6}, \bibinfo{pages}{1383--1394}.
\newblock \DOIprefix\doi{10.5194/tc-6-1383-2012}.
\bibitem[{McPhee(1992)}]{mcphee1992turbulent}
\bibinfo{author}{McPhee, M.G.}, \bibinfo{year}{1992}.
\newblock \bibinfo{title}{Turbulent heat flux in the upper ocean under sea ice}.
\newblock \bibinfo{journal}{Journal of Geophysical Research: Oceans} \bibinfo{volume}{97}, \bibinfo{pages}{5365--5379}.
\newblock \DOIprefix\doi{doi.org/10.1029/92JC00239}.
\bibitem[{Meier(2005)}]{Meier2005}
\bibinfo{author}{Meier, W.}, \bibinfo{year}{2005}.
\newblock \bibinfo{title}{Comparison of passive microwave ice concentration algorithm retrievals with {AVHRR} imagery in arctic peripheral seas}.
\newblock \bibinfo{journal}{IEEE Transactions on Geoscience and Remote Sensing} \bibinfo{volume}{43}, \bibinfo{pages}{1324--1337}.
\newblock \DOIprefix\doi{10.1109/TGRS.2005.846151}.
\bibitem[{Meier et~al.(2021)Meier, Fetterer, Windnagel and Stewart}]{Meier2021}
\bibinfo{author}{Meier, W.N.}, \bibinfo{author}{Fetterer, F.}, \bibinfo{author}{Windnagel, A.K.}, \bibinfo{author}{Stewart, J.S.}, \bibinfo{year}{2021}.
\newblock \bibinfo{title}{{NOAA/NSIDC Climate Data Record of Passive Microwave Sea Ice Concentration, Version 4}}.
\newblock \DOIprefix\doi{10.7265/efmz-2t65}.
\bibitem[{Meier et~al.(2014)Meier, Hovelsrud, van Oort, Key, Kovacs, Michel, Haas, Granskog, Gerland, Perovich, Makshtas and Reist}]{Meier2014}
\bibinfo{author}{Meier, W.N.}, \bibinfo{author}{Hovelsrud, G.K.}, \bibinfo{author}{van Oort, B.E.}, \bibinfo{author}{Key, J.R.}, \bibinfo{author}{Kovacs, K.M.}, \bibinfo{author}{Michel, C.}, \bibinfo{author}{Haas, C.}, \bibinfo{author}{Granskog, M.A.}, \bibinfo{author}{Gerland, S.}, \bibinfo{author}{Perovich, D.K.}, \bibinfo{author}{Makshtas, A.}, \bibinfo{author}{Reist, J.D.}, \bibinfo{year}{2014}.
\newblock \bibinfo{title}{Arctic sea ice in transformation: A review of recent observed changes and impacts on biology and human activity}.
\newblock \bibinfo{journal}{Reviews of Geophysics} \bibinfo{volume}{52}, \bibinfo{pages}{185--217}.
\newblock \DOIprefix\doi{10.1002/2013RG000431}.
\bibitem[{Notz and Stroeve(2016)}]{Notz2016}
\bibinfo{author}{Notz, D.}, \bibinfo{author}{Stroeve, J.}, \bibinfo{year}{2016}.
\newblock \bibinfo{title}{{Observed Arctic sea-ice loss directly follows anthropogenic CO2 emission}}.
\newblock \bibinfo{journal}{Science} \bibinfo{volume}{354}, \bibinfo{pages}{747--750}.
\newblock \DOIprefix\doi{10.1126/science.aag2345}.
\bibitem[{Overland and Wang(2013)}]{Overland2013}
\bibinfo{author}{Overland, J.E.}, \bibinfo{author}{Wang, M.}, \bibinfo{year}{2013}.
\newblock \bibinfo{title}{When will the summer {Arctic} be nearly sea ice free?}
\newblock \bibinfo{journal}{Geophysical Research Letters} \bibinfo{volume}{40}, \bibinfo{pages}{2097--2101}.
\newblock \DOIprefix\doi{10.1002/grl.50316}.
\bibitem[{Palerme et~al.(2024)Palerme, Lavergne, Rusin, Melsom, Brajard, Kvanum, Macdonald~S{\o}rensen, Bertino and M\"uller}]{Palerme2024}
\bibinfo{author}{Palerme, C.}, \bibinfo{author}{Lavergne, T.}, \bibinfo{author}{Rusin, J.}, \bibinfo{author}{Melsom, A.}, \bibinfo{author}{Brajard, J.}, \bibinfo{author}{Kvanum, A.F.}, \bibinfo{author}{Macdonald~S{\o}rensen, A.}, \bibinfo{author}{Bertino, L.}, \bibinfo{author}{M\"uller, M.}, \bibinfo{year}{2024}.
\newblock \bibinfo{title}{Improving short-term sea ice concentration forecasts using deep learning}.
\newblock \bibinfo{journal}{The Cryosphere} \bibinfo{volume}{18}, \bibinfo{pages}{2161--2176}.
\newblock \DOIprefix\doi{10.5194/tc-18-2161-2024}.
\bibitem[{Palerme and Muller(2021)}]{Palerme2021}
\bibinfo{author}{Palerme, C.}, \bibinfo{author}{Muller, M.}, \bibinfo{year}{2021}.
\newblock \bibinfo{title}{Calibration of sea ice drift forecasts using random forest algorithms}.
\newblock \bibinfo{journal}{The Cryosphere} \bibinfo{volume}{15}, \bibinfo{pages}{3989--4004}.
\newblock \DOIprefix\doi{10.5194/tc-15-3989-2021}.
\bibitem[{Parkinson and Washington(1979)}]{Parkinson1979}
\bibinfo{author}{Parkinson, C.L.}, \bibinfo{author}{Washington, W.M.}, \bibinfo{year}{1979}.
\newblock \bibinfo{title}{A large-scale numerical model of sea ice}.
\newblock \bibinfo{journal}{Journal of Geophysical Research: Oceans} \bibinfo{volume}{84}, \bibinfo{pages}{311--337}.
\newblock \DOIprefix\doi{10.1029/JC084iC01p00311}.
\bibitem[{Petrou and Tian(2019)}]{Petrou2019}
\bibinfo{author}{Petrou, Z.I.}, \bibinfo{author}{Tian, Y.}, \bibinfo{year}{2019}.
\newblock \bibinfo{title}{Prediction of sea ice motion with convolutional long short-term memory networks}.
\newblock \bibinfo{journal}{IEEE Transactions on Geoscience and Remote Sensing} \bibinfo{volume}{57}, \bibinfo{pages}{6865--6876}.
\newblock \DOIprefix\doi{10.1109/TGRS.2019.2909057}.
\bibitem[{Raissi et~al.(2019)Raissi, Perdikaris and Karniadakis}]{Rassi2019_PINN}
\bibinfo{author}{Raissi, M.}, \bibinfo{author}{Perdikaris, P.}, \bibinfo{author}{Karniadakis, G.}, \bibinfo{year}{2019}.
\newblock \bibinfo{title}{Physics-informed neural networks: A deep learning framework for solving forward and inverse problems involving nonlinear partial differential equations}.
\newblock \bibinfo{journal}{Journal of Computational Physics} \bibinfo{volume}{378}, \bibinfo{pages}{686--707}.
\newblock \DOIprefix\doi{10.1016/j.jcp.2018.10.045}.
\bibitem[{Rampal et~al.(2016)Rampal, Bouillon, \'Olason and Morlighem}]{netXtSIM2016}
\bibinfo{author}{Rampal, P.}, \bibinfo{author}{Bouillon, S.}, \bibinfo{author}{\'Olason, E.}, \bibinfo{author}{Morlighem, M.}, \bibinfo{year}{2016}.
\newblock \bibinfo{title}{{neXtSIM: a new Lagrangian sea ice model}}.
\newblock \bibinfo{journal}{The Cryosphere} \bibinfo{volume}{10}, \bibinfo{pages}{1055--1073}.
\newblock \DOIprefix\doi{10.5194/tc-10-1055-2016}.
\bibitem[{Ren and Li(2021)}]{Ren2021}
\bibinfo{author}{Ren, Y.}, \bibinfo{author}{Li, X.}, \bibinfo{year}{2021}.
\newblock \bibinfo{title}{{Predicting Daily Arctic Sea Ice Concentration in the Melt Season Based on a Deep Fully Convolution Network Model}}, in: \bibinfo{booktitle}{2021 IEEE International Geoscience and Remote Sensing Symposium IGARSS}, pp. \bibinfo{pages}{5540--5543}.
\newblock \DOIprefix\doi{10.1109/IGARSS47720.2021.9554118}.
\bibitem[{Ren et~al.(2022)Ren, Li and Zhang}]{Ren2022}
\bibinfo{author}{Ren, Y.}, \bibinfo{author}{Li, X.}, \bibinfo{author}{Zhang, W.}, \bibinfo{year}{2022}.
\newblock \bibinfo{title}{{A Data-Driven Deep Learning Model for Weekly Sea Ice Concentration Prediction of the Pan-Arctic During the Melting Season}}.
\newblock \bibinfo{journal}{IEEE Transactions on Geoscience and Remote Sensing} \bibinfo{volume}{60}, \bibinfo{pages}{1--19}.
\newblock \DOIprefix\doi{10.1109/TGRS.2022.3177600}.
\bibitem[{Riel and Minchew(2023)}]{Riel2023}
\bibinfo{author}{Riel, B.}, \bibinfo{author}{Minchew, B.}, \bibinfo{year}{2023}.
\newblock \bibinfo{title}{Variational inference of ice shelf rheology with physics-informed machine learning}.
\newblock \bibinfo{journal}{Journal of Glaciology} \bibinfo{volume}{69}, \bibinfo{pages}{1167–1186}.
\newblock \DOIprefix\doi{10.1017/jog.2023.8}.
\bibitem[{Riel et~al.(2021)Riel, Minchew and Bischoff}]{Riel2021}
\bibinfo{author}{Riel, B.}, \bibinfo{author}{Minchew, B.}, \bibinfo{author}{Bischoff, T.}, \bibinfo{year}{2021}.
\newblock \bibinfo{title}{Data-driven inference of the mechanics of slip along glacier beds using physics-informed neural networks: Case study on {Rutford Ice Stream, Antarctica}}.
\newblock \bibinfo{journal}{Journal of Advances in Modeling Earth Systems} \bibinfo{volume}{13}, \bibinfo{pages}{e2021MS002621}.
\newblock \DOIprefix\doi{10.1029/2021MS002621}.
\bibitem[{Ronneberger et~al.(2015)Ronneberger, Fischer and Brox}]{Ronneberger2015}
\bibinfo{author}{Ronneberger, O.}, \bibinfo{author}{Fischer, P.}, \bibinfo{author}{Brox, T.}, \bibinfo{year}{2015}.
\newblock \bibinfo{title}{{U-Net}: Convolutional networks for biomedical image segmentation}, in: \bibinfo{editor}{Navab, N.}, \bibinfo{editor}{Hornegger, J.}, \bibinfo{editor}{Wells, W.M.}, \bibinfo{editor}{Frangi, A.F.} (Eds.), \bibinfo{booktitle}{Medical Image Computing and Computer-Assisted Intervention -- MICCAI 2015}, \bibinfo{publisher}{Springer International Publishing}, \bibinfo{address}{Cham}. pp. \bibinfo{pages}{234--241}.
\bibitem[{Sadeghi et~al.(2020)Sadeghi, Nguyen, Hsu and Sorooshian}]{SADEGHI2020_EMS_CNN}
\bibinfo{author}{Sadeghi, M.}, \bibinfo{author}{Nguyen, P.}, \bibinfo{author}{Hsu, K.}, \bibinfo{author}{Sorooshian, S.}, \bibinfo{year}{2020}.
\newblock \bibinfo{title}{Improving near real-time precipitation estimation using a {U-Net} convolutional neural network and geographical information}.
\newblock \bibinfo{journal}{Environmental Modelling \& Software} \bibinfo{volume}{134}, \bibinfo{pages}{104856}.
\newblock \DOIprefix\doi{10.1016/j.envsoft.2020.104856}.
\bibitem[{{Salas Mélia}(2002)}]{SALASMELIA2002}
\bibinfo{author}{{Salas Mélia}, D.}, \bibinfo{year}{2002}.
\newblock \bibinfo{title}{A global coupled sea ice–ocean model}.
\newblock \bibinfo{journal}{Ocean Modelling} \bibinfo{volume}{4}, \bibinfo{pages}{137--172}.
\newblock \DOIprefix\doi{10.1016/S1463-5003(01)00015-4}.
\bibitem[{Satorras et~al.(2022)Satorras, Hoogeboom and Welling}]{Satorras2022_egcn}
\bibinfo{author}{Satorras, V.G.}, \bibinfo{author}{Hoogeboom, E.}, \bibinfo{author}{Welling, M.}, \bibinfo{year}{2022}.
\newblock \bibinfo{title}{E(n) equivariant graph neural networks}.
\newblock \DOIprefix\doi{10.48550/arXiv.2102.09844}, \href{http://arxiv.org/abs/2102.09844}{{\tt arXiv:2102.09844}}.
\bibitem[{Schütt et~al.(2017)Schütt, Kindermans, Sauceda, Chmiela, Tkatchenko and Müller}]{Schutt_2017_Schnet}
\bibinfo{author}{Schütt, K.T.}, \bibinfo{author}{Kindermans, P.J.}, \bibinfo{author}{Sauceda, H.E.}, \bibinfo{author}{Chmiela, S.}, \bibinfo{author}{Tkatchenko, A.}, \bibinfo{author}{Müller, K.R.}, \bibinfo{year}{2017}.
\newblock \bibinfo{title}{{SchNet}: A continuous-filter convolutional neural network for modeling quantum interactions}.
\newblock \DOIprefix\doi{10.48550/arXiv.1706.08566}, \href{http://arxiv.org/abs/1706.08566}{{\tt arXiv:1706.08566}}.
\bibitem[{Stark et~al.(2008)Stark, Ridley, Martin and Hines}]{Stark2008}
\bibinfo{author}{Stark, J.D.}, \bibinfo{author}{Ridley, J.}, \bibinfo{author}{Martin, M.}, \bibinfo{author}{Hines, A.}, \bibinfo{year}{2008}.
\newblock \bibinfo{title}{Sea ice concentration and motion assimilation in a sea ice-ocean model}.
\newblock \bibinfo{journal}{Journal of Geophysical Research: Oceans} \bibinfo{volume}{113}, \bibinfo{pages}{1--19}.
\newblock \DOIprefix\doi{10.1029/2007JC004224}.
\bibitem[{Stroeve and Notz(2018)}]{Stroeve2018}
\bibinfo{author}{Stroeve, J.}, \bibinfo{author}{Notz, D.}, \bibinfo{year}{2018}.
\newblock \bibinfo{title}{Changing state of {Arctic} sea ice across all seasons}.
\newblock \bibinfo{journal}{Environmental Research Letters} \bibinfo{volume}{13}, \bibinfo{pages}{103001}.
\newblock \DOIprefix\doi{10.1088/1748-9326/aade56}.
\bibitem[{Teisberg et~al.(2021)Teisberg, Schroeder and MacKie}]{Teisberg2021}
\bibinfo{author}{Teisberg, T.O.}, \bibinfo{author}{Schroeder, D.M.}, \bibinfo{author}{MacKie, E.J.}, \bibinfo{year}{2021}.
\newblock \bibinfo{title}{A machine learning approach to mass-conserving ice thickness interpolation}, in: \bibinfo{booktitle}{2021 IEEE International Geoscience and Remote Sensing Symposium IGARSS}, pp. \bibinfo{pages}{8664--8667}.
\newblock \DOIprefix\doi{10.1109/IGARSS47720.2021.9555002}.
\bibitem[{Thorndike and Colony(1982)}]{Thorndike1982}
\bibinfo{author}{Thorndike, A.S.}, \bibinfo{author}{Colony, R.}, \bibinfo{year}{1982}.
\newblock \bibinfo{title}{Sea ice motion in response to geostrophic winds}.
\newblock \bibinfo{journal}{Journal of Geophysical Research: Oceans} \bibinfo{volume}{87}, \bibinfo{pages}{5845--5852}.
\newblock \DOIprefix\doi{10.1029/JC087iC08p05845}.
\bibitem[{Timmermann et~al.(2009)Timmermann, Danilov, Schröter, Böning, Sidorenko and Rollenhagen}]{TIMMERMANN2009}
\bibinfo{author}{Timmermann, R.}, \bibinfo{author}{Danilov, S.}, \bibinfo{author}{Schröter, J.}, \bibinfo{author}{Böning, C.}, \bibinfo{author}{Sidorenko, D.}, \bibinfo{author}{Rollenhagen, K.}, \bibinfo{year}{2009}.
\newblock \bibinfo{title}{Ocean circulation and sea ice distribution in a finite element global sea ice–ocean model}.
\newblock \bibinfo{journal}{Ocean Modelling} \bibinfo{volume}{27}, \bibinfo{pages}{114--129}.
\newblock \DOIprefix\doi{10.1016/j.ocemod.2008.10.009}.
\bibitem[{Tschudi et~al.(2019)Tschudi, Meier, Fowler and Maslanik}]{NSIDC}
\bibinfo{author}{Tschudi, M.}, \bibinfo{author}{Meier, W. N.~Stewart, J.S.}, \bibinfo{author}{Fowler, C.}, \bibinfo{author}{Maslanik, J.}, \bibinfo{year}{2019}.
\newblock \bibinfo{title}{{Polar Pathfinder Daily 25 km EASE-Grid Sea Ice Motion Vectors, Version 4}}.
\newblock \DOIprefix\doi{10.5067/INAWUWO7QH7B}.
\bibitem[{Tschudi et~al.(2020)Tschudi, Meier and Stewart}]{Tschudi2020}
\bibinfo{author}{Tschudi, M.A.}, \bibinfo{author}{Meier, W.N.}, \bibinfo{author}{Stewart, J.S.}, \bibinfo{year}{2020}.
\newblock \bibinfo{title}{{An enhancement to sea ice motion and age products at the National Snow and Ice Data Center (NSIDC)}}.
\newblock \bibinfo{journal}{The Cryosphere} \bibinfo{volume}{14}, \bibinfo{pages}{1519--1536}.
\newblock \DOIprefix\doi{10.5194/tc-14-1519-2020}.
\bibitem[{Wang et~al.(2014)Wang, Danilov, Sidorenko, Timmermann, Wekerle, Wang, Jung and Schr\"oter}]{FESOM2014}
\bibinfo{author}{Wang, Q.}, \bibinfo{author}{Danilov, S.}, \bibinfo{author}{Sidorenko, D.}, \bibinfo{author}{Timmermann, R.}, \bibinfo{author}{Wekerle, C.}, \bibinfo{author}{Wang, X.}, \bibinfo{author}{Jung, T.}, \bibinfo{author}{Schr\"oter, J.}, \bibinfo{year}{2014}.
\newblock \bibinfo{title}{{The Finite Element Sea Ice-Ocean Model (FESOM) v.1.4: formulation of an ocean general circulation model}}.
\newblock \bibinfo{journal}{Geoscientific Model Development} \bibinfo{volume}{7}, \bibinfo{pages}{663--693}.
\newblock \DOIprefix\doi{10.5194/gmd-7-663-2014}.
\bibitem[{Wang et~al.(2022)Wang, Liu, Key and Dworak}]{Wang2022_new}
\bibinfo{author}{Wang, X.}, \bibinfo{author}{Liu, Y.}, \bibinfo{author}{Key, J.R.}, \bibinfo{author}{Dworak, R.}, \bibinfo{year}{2022}.
\newblock \bibinfo{title}{A new perspective on four decades of changes in {Arctic} sea ice from satellite observations}.
\newblock \bibinfo{journal}{Remote Sensing} \bibinfo{volume}{14}.
\newblock \DOIprefix\doi{10.3390/rs14081846}.
\bibitem[{Wang et~al.(2024)Wang, Shen, Salahshour, Cetin, Iftekharuddin, Tahvildari, Huang, Harris, Ampofo and Goodall}]{WANG2024_EMS_CNN}
\bibinfo{author}{Wang, Y.}, \bibinfo{author}{Shen, Y.}, \bibinfo{author}{Salahshour, B.}, \bibinfo{author}{Cetin, M.}, \bibinfo{author}{Iftekharuddin, K.}, \bibinfo{author}{Tahvildari, N.}, \bibinfo{author}{Huang, G.}, \bibinfo{author}{Harris, D.K.}, \bibinfo{author}{Ampofo, K.}, \bibinfo{author}{Goodall, J.L.}, \bibinfo{year}{2024}.
\newblock \bibinfo{title}{Urban flood extent segmentation and evaluation from real-world surveillance camera images using deep convolutional neural network}.
\newblock \bibinfo{journal}{Environmental Modelling \& Software} \bibinfo{volume}{173}, \bibinfo{pages}{105939}.
\newblock \DOIprefix\doi{10.1016/j.envsoft.2023.105939}.
\bibitem[{Woo et~al.(2018)Woo, Park, Lee and Kweon}]{Woo2018}
\bibinfo{author}{Woo, S.}, \bibinfo{author}{Park, J.}, \bibinfo{author}{Lee, J.Y.}, \bibinfo{author}{Kweon, I.S.}, \bibinfo{year}{2018}.
\newblock \bibinfo{title}{{CBAM: Convolutional Block Attention Module}}, in: \bibinfo{editor}{Ferrari, V.}, \bibinfo{editor}{Hebert, M.}, \bibinfo{editor}{Sminchisescu, C.}, \bibinfo{editor}{Weiss, Y.} (Eds.), \bibinfo{booktitle}{Computer Vision -- ECCV 2018}, \bibinfo{publisher}{Springer International Publishing}, \bibinfo{address}{Cham}. pp. \bibinfo{pages}{3--19}.
\newblock \DOIprefix\doi{10.1007/978-3-030-01234-2\_1}.
\bibitem[{Yan and Huang(2018)}]{Yan2018}
\bibinfo{author}{Yan, Q.}, \bibinfo{author}{Huang, W.}, \bibinfo{year}{2018}.
\newblock \bibinfo{title}{Sea ice sensing from {GNSS-R} data using convolutional neural networks}.
\newblock \bibinfo{journal}{IEEE Geoscience and Remote Sensing Letters} \bibinfo{volume}{15}, \bibinfo{pages}{1510--1514}.
\newblock \DOIprefix\doi{10.1109/LGRS.2018.2852143}.
\bibitem[{Zhai and Bitz(2021)}]{zhai2021}
\bibinfo{author}{Zhai, J.}, \bibinfo{author}{Bitz, C.M.}, \bibinfo{year}{2021}.
\newblock \bibinfo{title}{A machine learning model of {Arctic} sea ice motions}.
\newblock \DOIprefix\doi{10.48550/arXiv.2108.10925}, \href{http://arxiv.org/abs/2108.10925}{{\tt arXiv:2108.10925}}.
\bibitem[{Årthun et~al.(2021)Årthun, Onarheim, Dörr and Eldevik}]{Årthun2021}
\bibinfo{author}{Årthun, M.}, \bibinfo{author}{Onarheim, I.H.}, \bibinfo{author}{Dörr, J.}, \bibinfo{author}{Eldevik, T.}, \bibinfo{year}{2021}.
\newblock \bibinfo{title}{{The Seasonal and Regional Transition to an Ice-Free Arctic}}.
\newblock \bibinfo{journal}{Geophysical Research Letters} \bibinfo{volume}{48}, \bibinfo{pages}{e2020GL090825}.
\newblock \DOIprefix\doi{10.1029/2020GL090825}.

\end{thebibliography}






\end{document}